\documentclass{article}
\usepackage{makecell}

\usepackage[square,sort,comma]{natbib}

\usepackage{microtype}
\usepackage{graphicx}
\usepackage{adjustbox}   
\usepackage{tabularx}

\usepackage{pgfplots}
\pgfplotsset{compat=1.18} 

\usepackage{subcaption}
\DeclareCaptionLabelFormat{subfig}{Figure~\thefigure~(#2)}
\usepackage{booktabs} 
\newcommand{\ours}{EvoMU}
\def\redc{\cellcolor[HTML]{FF999A}}
\def\orangec{\cellcolor[HTML]{FFCC99}}
\def\yellowc{\cellcolor[HTML]{FFF8AD}}
\usepackage[most]{tcolorbox} 
\usepackage{standalone}
\usepackage[table,dvipsnames]{xcolor}

\usepackage[table]{xcolor}
\definecolor{lightyellow}{RGB}{255, 255, 224} 

\usepackage{enumitem}
\setlist[itemize]{
    leftmargin=15pt,      
    itemsep=2pt,          
    parsep=0pt,           
    topsep=0pt,        
    partopsep=0pt      
}
\setlist{leftmargin=*, itemsep=0pt}
\setlist[itemize,1]{leftmargin=10pt}  
\setlist[itemize,2]{leftmargin=15pt}  

\usepackage{hyperref}

\usepackage[preprint]{icml2026}

\usepackage{amsmath}
\usepackage{amssymb}
\usepackage{mathtools}
\usepackage{amsthm}

\newenvironment{promptbox}[2]{%
  \begin{tcolorbox}[
    enhanced,
    breakable,
    colback=#1!8!white,
    colframe=#1!70!black,
    drop shadow,
    arc=2mm, sharp corners=south,
    fontupper=\scriptsize,
    coltitle=black,
    varwidth boxed title*,
    boxed title style={boxrule=0.4pt, colback=white, sharp corners},
    attach boxed title to top left={yshift=-\tcboxedtitleheight/2, xshift=2mm},
    title={\bfseries\scriptsize #2},
  ]}{\end{tcolorbox}}

\usepackage[capitalize,noabbrev]{cleveref}

\theoremstyle{plain}

\theoremstyle{definition}

\theoremstyle{remark}

\usepackage{multirow}

\usepackage[textsize=tiny]{todonotes}

\icmltitlerunning{EvoMU: Evolutionary Machine Unlearning}

\begin{document}

\twocolumn[
  \icmltitle{EvoMU: Evolutionary Machine Unlearning}



  \icmlsetsymbol{equal}{*}

  \begin{icmlauthorlist}
    \icmlauthor{Paweł Batorski}{hhu}
    \icmlauthor{Paul Swoboda}{hhu}
  \end{icmlauthorlist}

  \icmlaffiliation{hhu}{Heinrich Heine Universität Düsseldorf, Germany}

  \icmlcorrespondingauthor{Pawel Batorski}{pawel.batorski@hhu.de}

  \icmlkeywords{Machine Learning, ICML}

  \vskip 0.3in
]

\begin{figure*}
  \centering
  \includegraphics[width=\linewidth]{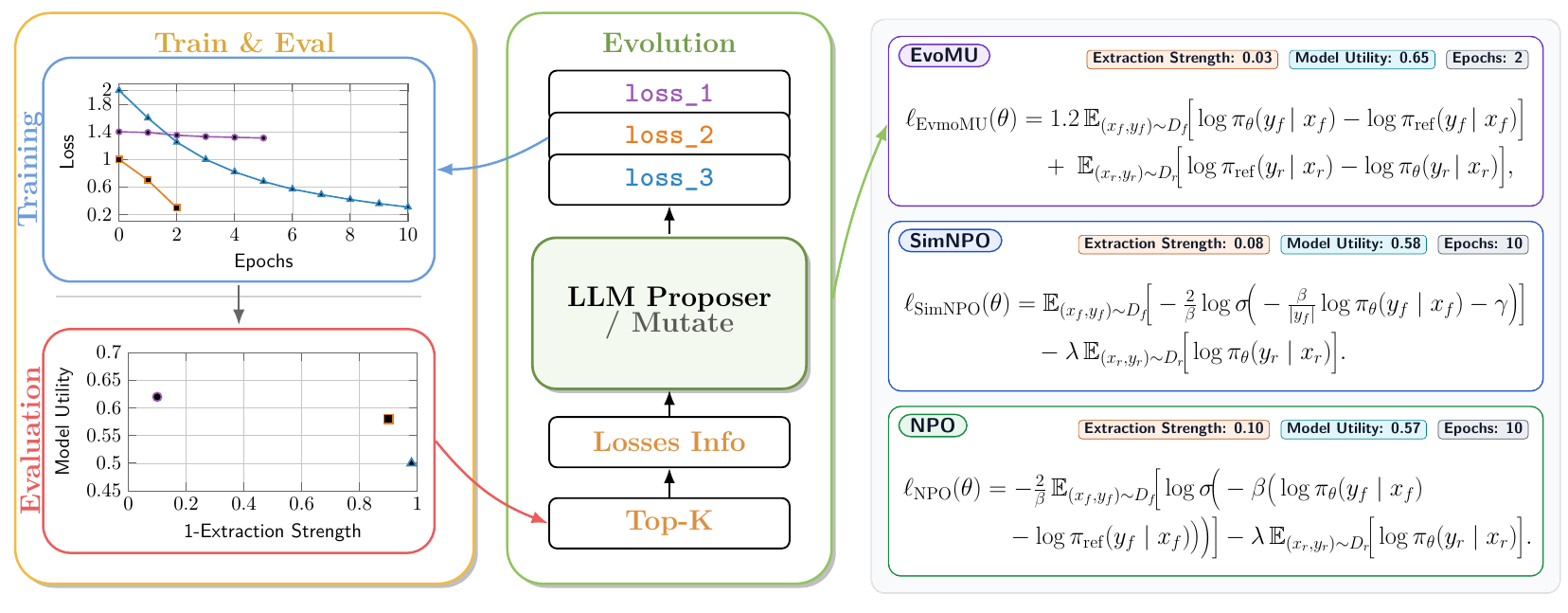}
  \caption{
     High level overview of \ours{}. A LLM proposes candidate unlearning loss functions along with their training budget. For each candidate, we fine-tune the base model with LoRA adapters on forget and retain data, then evaluate the resulting checkpoint using standard forgetting and utility metrics. We select the top-$K$ objectives and summarize their full feedback (loss code, training curves, and evaluation scores). Conditioned on this feedback, the LLM mutates the objectives by adjusting coefficients, modifying structure, and optionally changing the training budget, to produce the next generation. Repeating this verify--select--mutate loop yields task-specific unlearning losses that match or outperform human-designed ones.
}
  \label{fig:teaser}
\end{figure*}



\printAffiliationsAndNotice{}  

\begin{abstract}
Machine unlearning aims to unlearn specified training data (e.g.\ sensitive or copyrighted material). A prominent approach is to fine-tune an existing model with an unlearning loss that retains overall utility.
The space of suitable unlearning loss functions is vast, making the search for an optimal loss function daunting.
Additionally, there might not even exist a universally optimal loss function: differences in the structure and overlap of the forget and retain data can cause a loss to work well in one setting but over-unlearn or under-unlearn in another.
Our approach \ours{} tackles these two challenges simultaneously. An evolutionary search procedure automatically finds task-specific losses in the vast space of possible unlearning loss functions.
This allows us to find dataset-specific losses that match or outperform existing losses from the literature, without the need for a human-in-the-loop. 
This work is therefore an instance of automatic scientific discovery, a.k.a.\ an AI co-scientist.
In contrast to previous AI co-scientist works, we do so on a budget: 
We achieve SotA results using a small 4B parameter model (Qwen3-4B-Thinking), showing the potential of AI co-scientists with limited computational resources.
Our experimental evaluation shows that we surpass previous loss-based unlearning formulations on TOFU-$5\%$, TOFU-$10\%$, MUSE and WMDP by synthesizing novel unlearning losses.
Our code is available at \url{https://github.com/Batorskq/EvoMU}  

%
\end{abstract}

\section{Introduction}

As large language models (LLMs) grow in capability, they also increasingly memorize and reproduce sensitive, harmful, or copyrighted content \citep{huang2024trustllm, carlini2022quantifying, wang2023decodingtrust, wei2024evaluating, ousidhoum2021probing, barbulescu2024each, dou2025copyright}. In many real deployments, this creates a practical need for \emph{machine unlearning}: removing the influence of specified training data from a trained model without retraining from scratch \citep{cao2015machineunlearning, ginart2019deletion, bourtoule2021sisa}. Beyond its origins in data deletion and privacy regulations (e.g., the ``right to be forgotten'' \citep{rosen2011right, hoofnagle2019european}), unlearning is now motivated by content compliance, safety, and intellectual property concerns \citep{liu2025rethinking}. 

A dominant approach for unlearning in modern LLMs is post-training fine-tuning with unlearning losses that suppress unlearning data while preserving general utility.
They include gradient-ascent or gradient-difference formulations \citep{yao2024llmunlearning, liu2022continual, maini2024tofu}, preference-optimization objectives that discourage undesired completions \citep{npo, simnpo, mekala2025altpo}, and distillation or logit/representation-regularization schemes \citep{rkld, undial, li2024wmdp}.
However, loss design remains a major bottleneck.
Hand-designed losses may be brittle: small changes in coefficients, margins, or reference signals can shift the trade-off between effective forgetting and over-unlearning.
%
Additionally, unlearning losses are rarely universal.
The difficulty of selective forgetting depends strongly on the structure of the unlearning instance: how concentrated or diffuse the target content is, how much the forget and retain distributions overlap, and whether the forget targets are isolated facts, stylistic patterns, or long-form passages.
This structure changes the dominant failure mode.
When forget and retain share entities, phrasing, or underlying facts, a naive loss may suppress too broad regions of behavior (e.g., refusing or becoming uncertain in-domain) rather than selectively removing the intended information.
What matters is not only ``how much forgetting'' is applied, but how the loss balances forget vs.\ retain and how it prevents degenerate solutions.
These considerations suggest that loss design should adapt to the data geometry of the unlearning problem rather than rely on a single fixed loss.

The above observation, namely large and problem-dependant, yet well-specified search space with clear evaluation protocol, makes machine unlearning a good target for automated scientific discovery (ASD).
Modern ASD has seen great success 
(i)~when using LLMs for code generation, experimentation and iterative refinement \citep{wang2023scientific, lu2024ai, schmidgall2025agent, jansen2025codescientist, yamada2025ai} and
(ii)~when candidates are executable and progress is measured by automatic metrics, enable efficient, auditable search loops (e.g., algorithm or reward-function discovery) \citep{fawzi2022alphatensor, ma2023eureka, richardson2024understanding, cheng2025language}.
%
In this work, we study unlearning loss design as an ASD problem and introduce \ours{}, an evolutionary objective-discovery framework.
\ours{} uses a code-generating LLM to propose candidate unlearning loss, fine-tunes lightweight LoRA adapters for each candidate, evaluates with standardized unlearning metrics, and iteratively refines the best loss using an evolutionary search.
Crucially, we show that this discovery loop does not require frontier-scale or closed-weight models: a small open model (Qwen3-4B-Thinking) is sufficient to synthesize novel, effective losses that match or outperform strong human-designed baselines across multiple benchmarks.

Our contributions are:
\begin{itemize}
    \item \textbf{ASD for unlearning.} We formulate unlearning loss design as an automated discovery problem and present an end-to-end pipeline that proposes, trains, evaluates, and refines executable loss functions.
    \item \textbf{Small-model effectiveness.} We demonstrate that a 4B-parameter open LLM can effectively drive loss discovery, contrasting with prior ASD systems that rely on very large or closed models.
    \item \textbf{Task-specific losses matter.}
    Across TOFU, MUSE, and WMDP, the best-performing losses differ substantially, indicating that unlearning losses should be tailored to the forget/retain data structure rather than treated as universal. 
    \item \textbf{LLM-sampled losses are a strong baseline.}
    Even without refinement, randomly sampled candidate losses can be competitive with established losses (e.g., NPO), highlighting that loss choice is a major lever.
    \item \textbf{Empirical structure of effective losses.} Across benchmarks, many top-performing discovered losses are simple and often do not require explicit reference-model terms, suggesting underexplored regions of the unlearning loss space.
\end{itemize}

\section{Related Work}

\subsection{Machine Unlearning for LLMs.}
Recent surveys and position papers argue that unlearning for LLMs should be viewed as a continuum of intervention surfaces, spanning inference-time suppression, parameter editing, and weight-level updates, while emphasizing that reliable evaluation and meaningful guarantees remain open problems \citep{liu2025rethinking, ren2025sok, thaker2025weakbenchmarks}. Broadly, existing approaches fall into three (overlapping) families: (i) mitigation mechanisms that restrict access to unwanted content without directly optimizing a forgetting loss, (ii) localized editing or model-merging methods that overwrite specific associations, and (iii) training-time unlearning that updates model parameters (often via parameter-efficient fine-tuning) under a designed loss.

\paragraph{Mitigation without optimizing a forget loss.}
This approach limits access to knowledge to be unlearned without updating the base model.
This includes inference-time controls such as decoding- or logit-level interventions and related approximate-unlearning techniques \citep{eldan2023s, yu2021differentially, huang2024offset, ji2024reversing}, as well as prompt- and context-based defenses that steer generation away from target content \citep{liu2024large, thaker2024guardrail, pawelczyk2023context}. These approaches are often lightweight and deployment-friendly, but they may deflect rather than remove information, and auxiliary mechanisms can themselves become new loci of leakage if they encode or expose the content slated for deletion.
Currently, such techniques seem to be less capable than unlearning loss based ones.

\paragraph{Weight-level interventions and parameter editing.}
This approach performs localized parameter editing or model merging to overwrite specific associations \citep{ilharco2022editing, chen2023unlearn, barbulescu2024each, meng2022rome, meng2023memit}. Such methods can provide fine-grained control and fast updates, and they are increasingly used as primitives for targeted behavioral edits. However, editing typically targets a narrow mechanism (e.g., a small set of facts/associations) and may not scale cleanly to distributional “forget sets,” where the loss is to suppress a broader region of behaviors rather than a single mapping.
Parameter editing techniques perform worse than unlearning loss approaches.

\paragraph{Training-time unlearning via parameter updates and loss design.}
The currently best performing unlearning approaches perform unlearning by updating model parameters via additional training, frequently with parameter-efficient fine-tuning, to reduce the likelihood of producing forget-set targets while preserving performance on retain data.
This family includes early privacy-motivated formulations \citep{jang2023privacyunlearn} and many loss designs: KL-regularized or distillation-based variants \citep{rkld, vasilev2025unilogit, undial}, ``I don't know'' targets on forget prompts \citep{maini2024tofu}, counterfactual fine-tuning that selects alternative answers using memorization signals \citep{gu2024meow}, and loss shaping schemes that operate even with forget-only data \citep{flat}. Classic likelihood-based primitives such as gradient ascent on the forget set (and related gradient-difference formulations) remain strong baselines and are frequently used as building blocks \citep{ga, yao2024llmunlearning, maini2024tofu, barbulescu2024each, liu2022continual}. More recent work explores preference-style losses inspired by alignment (DPO/NPO and simplifications) \citep{dpo, npo, simnpo, mekala2025altpo}, stronger optimization procedures \citep{jia2024soul}, and principled analyses/unifications of loss families \citep{wang2025geffect} as well as methods that explicitly manage the retention/forgetting trade-off \citep{wang2025gru}.
In parallel, unlearning updates inherit the optimization sensitivity of modern fine-tuning. 
Recent studies and methods further investigate optimization choices and stabilization strategies for unlearning-style losses \citep{choi2025opt, foret2021sam, bhaila2025soft}.

\paragraph{Exact/certified unlearning and formal perspectives.}
Beyond empirical post-training updates, this line of work studies conditions under which deletion can be made exact or otherwise formally controlled, and when forgetting may be fundamentally difficult. Work on exact/fast deletion procedures and on limits of learnability/unlearnability provides complementary perspectives on what can be guaranteed (and what cannot) under practical constraints \citep{muresanu2024unlearnable, muresanu2025fastexact}. 

\paragraph{Robustness, reversibility, relearning, and evaluation brittleness.}
Across training-time methods, a recurring challenge is the tension between strong forgetting and over-unlearning, where utility degrades as forgetting pressure increases \citep{tian2024forget, yang2025exploring}. Multiple works further highlight that post-hoc unlearning can be reversible: benign fine-tuning may restore seemingly forgotten data, and unlearning can behave more like obfuscation than deletion under realistic threat models \citep{hu2025jogging, xu2025reversibility}.
Closely related, recent work studies relearning dynamics and how quickly forgotten data can return under subsequent training \citep{xu2025relearn}.
Robustness can also fail unexpectedly. For instance, downstream quantization can significantly undermine unlearning effects \citep{zhang2025quantfail}.
These concerns motivate practical/robust procedures and primitives \citep{xu2025obliviate, yu2025unierase} and continued scrutiny of evaluation protocols and privacy-leakage measurements \citep{rashid2025forget, chundawat2024fragile}.
More broadly, several recent efforts aim to better characterize what unlearning changes internally and how evaluation should be interpreted across regimes \citep{jin2024rwku, di2024dissect, wei2025llms, yuan2024closer, fan2025towards, cha2024towards, zhuang2025seuf, yuan2025towards, wang2025invariance}.
Our method is motivated by the practical implication of these findings: loss choice is often the dominant lever shaping failure modes (under-forgetting vs.\ over-unlearning), and thus loss design should adapt to the structure of the unlearning instance.

\begin{figure*}[ht]
  \centering
\includegraphics[width=\linewidth]{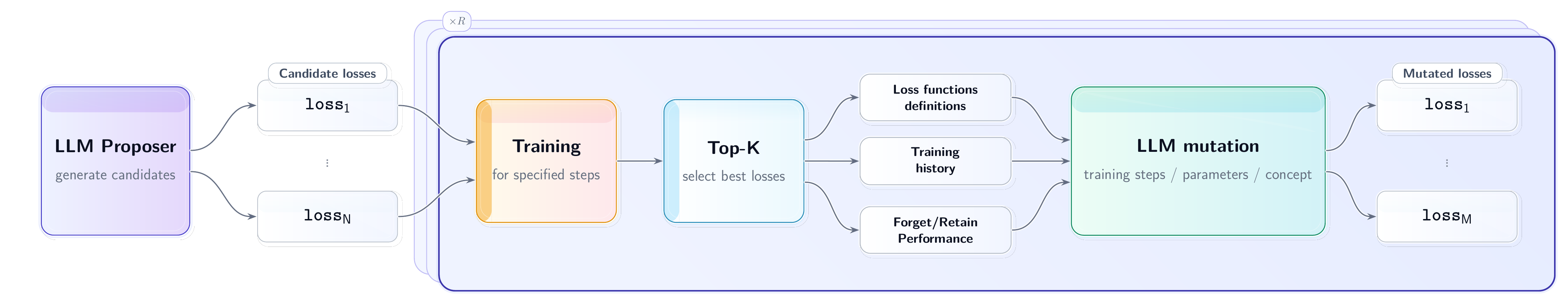}
  \caption{
  \ours{} overview. We use an LLM Proposer to generate an initial set of unlearning losses. Each loss is evaluated by training an LLM with it. The top-$K$ candidates are retained and mutated based on the obtained results. The mutation loop is repeated $R$ times.
  }
  \label{fig:method}
\end{figure*}

\subsection{AI in Scientific Discovery and verification-first search.}
The idea of automating parts of the scientific process long predates modern deep learning, with early systems focusing on symbolic search and heuristic-driven hypothesis formation and validation \citep{LangleySimonBradshawZytkow1987ScientificDiscovery}. More recently, AI-assisted discovery has broadened substantially, spanning literature-driven hypothesis generation, experimental planning, and closed-loop optimization pipelines \citep{wang2023scientific, liu2024aigs}. With the emergence of strong code-capable LLMs, several works have explored agentic or end-to-end discovery loops in which models propose ideas, implement experiments, and iteratively refine based on results \citep{lu2024ai, schmidgall2025agent, jansen2025codescientist, yamada2025ai}. A recurring limitation of open-ended automated science is that losses can be underspecified and evaluations difficult to verify automatically, which makes progress heavily dependent on expensive human feedback or brittle proxies.

ASD is easier to apply for well-specified discovery problems with automatic, reproducible evaluation, such as algorithm or loss design. For example, AlphaTensor demonstrates that reinforcement-learning-driven search can discover improved matrix multiplication algorithms under a precise scoring rule \citep{fawzi2022alphatensor}, while Eureka shows that LLMs can synthesize reward functions that effectively drive reinforcement-learning agents, again under clear downstream evaluation \citep{ma2023eureka}. Recent works include LLMs~\cite{cheng2025language} for neural network design and \citep{gottweis2025towards} for various scientific computing tasks. Our work falls in this category, but focuses on unlearning loss design. 
In contrast to prior ASD work that primarily relies on closed-source or very large LLMs, we show that a small 4B-parameter model, when paired with evolutionary search, can already perform effective automated scientific discovery.



\section{Method}
\label{sec:method}
We study machine unlearning as a loss design problem: given a forget set $\mathcal{D}_f$ and a retain set $\mathcal{D}_r$, we seek a training loss that reduces the model's likelihood of producing forbidden completions from $\mathcal{D}_f$ while preserving utility on $\mathcal{D}_r$.
Our approach \ours{} performs an evolutionary search over executable loss functions.
At each iteration, a code-generating LLM proposes candidate losses, we optimize each loss by fine-tuning with LoRA adapters, evaluate the resulting checkpoints using standardized unlearning metrics, and use the best-performing losses to seed the next round of refinements, mutating the best $K$ losses.

\paragraph{Problem Setup}
We study machine unlearning for an autoregressive language model $p_{\theta}$.
We are given a forget dataset (content that should be forgotten) $\mathcal{D}_f = \{(x^f_i,y^f_i) : i=1,\ldots,N_f\}$  and a retain dataset (content that should be preserved) $\mathcal{D}_r = \{(x^r_i,y^r_i) : i=1,\ldots,N_r\}$, where $x^f$ and $x^r$ are the forget and retain prompts and $y^f$ and $y^r$ are the respective responses.
Our goal is to update model parameters $\theta$ such that the model's behavior on $\mathcal{D}_f$ becomes less probable while utility on $\mathcal{D}_r$ is maintained.

For an input sequence $s=(x,y)$ (prompt $x$ and response $y$), we compute the average log-probability:
\begin{equation}
\label{eq:seqlogp}
\log p_{\theta}(y\mid x)
~\triangleq~
\frac{1}{|y|}
\sum_{t=1}^{|y|}
\log p_{\theta}(y_t \mid x, y_{<t})\,.
\end{equation}

We use the following average log-probs at each optimization step: 
$\mathbf{z}_f \in \mathbb{R}^{B}$ come from the forget batch and $\mathbf{z}_r \in \mathbb{R}^{B}$ from a retain batch evaluated by the current model.
Correspondingly $\mathbf{z}^{\mathrm{ref}}$ denote reference-model log-probabilities.
A candidate unlearning loss is a scalar function
$
\label{eq:genericloss}
\mathcal{L}(\mathbf{z}_f, \mathbf{z}_r, \mathbf{z}_f^{\mathrm{ref}}, \mathbf{z}_r^{\mathrm{ref}})\,.
$

\paragraph{Candidate loss space and constraints.}
We constrain our search space to be all python functions that take the four input parameters as in~\eqref{eq:genericloss} and return a single scalar value, implementing a differentiable pytorch method.
Additionally, in the doc-string we require the number of epochs the loss is optimized to be specified.
To keep the discovery setting verifiable and efficient, candidates may only use these statistics (no hidden states, gradients, sampling, or external calls).
Invalid candidates (exceptions, non-finite outputs, or missing required signature/docstring fields) are discarded.

\begin{table*}
\caption{Results on TOFU-5\% (LLaMA2-7B-Chat). Colors indicate rank (red: best, orange: second, yellow: third).}
\centering
\setlength{\tabcolsep}{3.5pt}
\renewcommand{\arraystretch}{1.15}
\begin{adjustbox}{max width=\textwidth}
\begin{tabular}{lcccccccccccc|c}
\toprule
& \multicolumn{3}{c}{Unlearning Efficacy} & \multicolumn{10}{c}{Utility Preservation} \\
\cmidrule(lr){2-4} \cmidrule(lr){5-14}
Method &
\multicolumn{3}{c}{Forget Set} &
\multicolumn{3}{c}{Real Authors} &
\multicolumn{3}{c}{World Facts} &
\multicolumn{3}{c|}{Retain Set} &
\multirow{2}{*}{\centering MU ($\uparrow$)}
\\
\cmidrule(lr){2-4} \cmidrule(lr){5-7} \cmidrule(lr){8-10} \cmidrule(lr){11-13}
& 1-Rouge-L$\uparrow$ & 1-Prob.$\uparrow$ & 1-Extr.\ Strength$\uparrow$ &
Rouge-L$\uparrow$ & Prob.$\uparrow$ & Truth ratio$\uparrow$ &
Rouge-L$\uparrow$ & Prob.$\uparrow$ & Truth ratio$\uparrow$ &
Rouge-L$\uparrow$ & Prob.$\uparrow$ & Truth ratio$\uparrow$ &
\\
\midrule
Original & 0.04 & 0.01 & 0.05 & 0.93 & 0.44 & 0.58 & 0.91 & 0.43 & 0.55 & 0.98 & 0.99 & 0.48 & 0.62 \\
Retain   & 0.61 & 0.85 & 0.93 & 0.92 & 0.44 & 0.57 & 0.90 & 0.43 & 0.54 & 0.97 & 0.99 & 0.48 & 0.62 \\ \hline

GradDiff & \redc 1.00 & \redc 1.00 & \orangec 0.96 & 0.59 & \redc 0.59 & \redc 0.81 & \yellowc 0.88 & 0.46 & \yellowc 0.59 & 0.42 & 0.49 & 0.48 & 0.56 \\
IDKDPO   & \orangec 0.98 & 0.40 & 0.85 & 0.65 & \yellowc 0.48 & 0.63 & 0.82 & 0.44 & 0.55 & \yellowc 0.55 & \orangec 0.86 & \orangec 0.57 & \yellowc 0.57 \\
RKLD     & 0.69 & 0.96 & \yellowc 0.92 & \redc 0.92 & 0.47 & 0.61 & 0.87 & \yellowc 0.47 & 0.58 & \orangec 0.58 & 0.52 & \yellowc 0.56 & 0.56 \\
NPO      & 0.73 & 0.94 & 0.90 & \orangec 0.91 & \orangec 0.50 & 0.62 & \redc 0.90 & \redc 0.50 & \redc 0.61 & 0.47 & 0.51 & \orangec 0.57 & \yellowc 0.57 \\
SimNPO   & \yellowc 0.74 & \yellowc 0.97 & \yellowc 0.92 & \yellowc 0.90 & \orangec 0.50 & \yellowc 0.64 & \redc 0.90 & \orangec 0.48 & \orangec 0.60 & 0.54 & \yellowc 0.56 & \redc 0.58 & \orangec 0.58 \\
\ours{}  & \redc 1.00 & \orangec 0.99 & \redc 0.97 & 0.89 & \orangec 0.50 & \orangec 0.65 & \orangec 0.89 & \yellowc 0.47 & \orangec 0.60 & \redc 0.90 & \redc 0.95 & 0.46 & \redc 0.65 \\ 

\bottomrule
\end{tabular}
\end{adjustbox}
\label{tab:tofu5}
\end{table*}

\paragraph{LLM for Proposing and Mutating Losses}
We use an LLM to initially propose candidate loss functions and number of iterations to optimize and then, in the evolutionary search phase, to mutate these.
For greater performance we use a thinking scratchpad, while the final output will contain python code for the produced loss function.

In order to encourage more diversity, we use a higher temperature for the thinking scratchpad.
To avoid errors in distilling the loss function from the scratchpad we use a lower temperature in this part.
We discard duplicates that were proposed by different invocations.
We also remove function candidates that are not executable.
Also, we try to repair functions that were not generated fully correctly, e.g.\: 
if a function returns multiple return values we average them to make the repaired function scalar-valued and
we replace numpy function calls by their pytorch equivalents.  The LLM Proposer prompt is included in Appendix \ref{sec:app_proposer} and exemplary initial losses can be found in Appendix~\ref{app:initial_losses}.

\paragraph{Training and Evaluation}
For each candidate loss $\mathcal{L}$, we fine-tune the base model using LoRA adapters~\cite{lora}, keeping the backbone weights frozen. Each loss function specifies its own training budget via an \texttt{epochs: iter} docstring, allowing the search to explore both losses and training duration. After training, we merge the learned LoRA weights into the base model to obtain a standalone checkpoint, and evaluate it with an external unlearning benchmark suite. This evaluation returns a number of metrics $m(\mathcal{L})$ that captures both forgetting efficacy on $\mathcal{D}_f$ and utility preservation on $\mathcal{D}_r$ (and additional held-out slices when available).

To rank candidates during search, we compute a scalar selection score
\begin{equation}
\label{eq:topkscore}
s(\mathcal{L}) ~=~ \tfrac{1}{2}\,\mathrm{Utility}(m(\mathcal{L})) ~+~ \tfrac{1}{2}\,\mathrm{Forget}(m(\mathcal{L})),
\end{equation}
where $\mathrm{Forget}(\cdot)$ averages the available normalized forgetting terms (e.g., $1{-}\mathrm{ROUGE}_f$ and $1{-}\mathrm{Prob}_f$) and $\mathrm{Utility}(\cdot)$ is a normalized utility summary returned by the evaluator. Any candidate that fails at generation, training, or evaluation is assigned zero score and is excluded from top-$K$ selection. 

\paragraph{Mutation}
At each mutation iteration, we select the top-$K$ candidates under $s(\cdot)$ as parents. For each parent, we provide the LLM with (i)~the loss code, (ii)~the per-epoch training loss history, (iii)~the chosen epoch budget, and (iv)~the full evaluation metrics. Conditioned on this feedback, the LLM proposes a set of mutated children losses, which may (a)~adjust coefficients, margins, or nonlinearities to better trade off forgetting and utility, (b)~introduce or remove reference-model terms, and/or (c)~modify the recommended number of epochs. 
We then repeat the same train/merge/evaluate procedure for the children and continue the evolutionary loop. The prompt for loss function refinement is included in Appendix \ref{sec:app_refine}.

\section{Experiments}

We fine-tune each base model using LoRA adapters with rank $r=8$ and scaling $\alpha=16$.
For each candidate loss proposed by \ours{}, the training budget is read directly from the loss function metadata: we use a variable number of epochs for TOFU and MUSE, and a variable number of optimization steps for WMDP. 

We evaluate \ours{} against established unlearning baselines on TOFU-5\% and TOFU-10\% \citep{maini2024tofu}, MUSE \citep{shi2024muse} and WMDP \citep{wmdp}.
Across all experiments, the LLM Proposer and Mutator is Qwen3-4B-Thinking \citep{yang2025qwen3}, while the unlearned (target) model depends on the benchmark.
All checkpoints are evaluated using the Open-Unlearning evaluation protocol \citep{dorna2025openunlearning}.
The specific forgetting and utility metrics vary by benchmark (as is standard in machine unlearning). A summary of the target models, tasks, and metrics for unlearning effectiveness and utility preservation is provided in Appendix~\ref{app:summary}.

All training runs are performed on two NVIDIA A100 GPUs (40GB).
For the evolutionary search, we sample $N=10$ initial candidate losses, select the top-$5$ according to the selection score, and generate $5$ refinements per parent (yielding $25$ candidates).
We then select the top-$3$ among these and generate $10$ additional refinements per parent. 
In total, our evolutionary search tests 65 losses in each run.
\ours{} runs for 10 to 20 hours depending on the benchmark.
We perform thinking stage decoding with temperature 0.6 and implementation with 0.2.


\begin{table*}
\caption{Results on TOFU-10\%. Colors indicate rank (red: best, orange: second, yellow: third). We unlearn Llama3.2-1B-Instruct.}
\centering
\setlength{\tabcolsep}{3.5pt}
\renewcommand{\arraystretch}{1.15}
\begin{adjustbox}{max width=0.95\textwidth}
\begin{tabular}{lcccccccccccc|c}
\toprule
& \multicolumn{3}{c}{Unlearning Efficacy} & \multicolumn{10}{c}{Utility Preservation} \\
\cmidrule(lr){2-4} \cmidrule(lr){5-14}
Method &
\multicolumn{3}{c}{Forget Set} &
\multicolumn{3}{c}{Real Authors} &
\multicolumn{3}{c}{World Facts} &
\multicolumn{3}{c|}{Retain Set} &
\multirow{2}{*}{\centering MU ($\uparrow$)}
\\
\cmidrule(lr){2-4} \cmidrule(lr){5-7} \cmidrule(lr){8-10} \cmidrule(lr){11-13}
& 1-Rouge-L$\uparrow$ & 1-Prob.$\uparrow$ & 1-Extr.\ Strength$\uparrow$ &
Rouge-L$\uparrow$ & Prob.$\uparrow$ & Truth ratio$\uparrow$ &
Rouge-L$\uparrow$ & Prob.$\uparrow$ & Truth ratio$\uparrow$ &
Rouge-L$\uparrow$ & Prob.$\uparrow$ & Truth ratio$\uparrow$
\\
\midrule
Original & 0.18 & 0.12 & 0.29 & 0.80 & 0.41 & 0.53 & 0.83 & 0.44 & 0.62 & 0.79 & 0.87 & 0.52 & 0.60 \\
Retain   & 0.62 & 0.88 & 0.94 & 0.83 & 0.39 & 0.50 & 0.80 & 0.43 & 0.62 & 0.83 & 0.88 & 0.51 & 0.59 \\ \hline

RMU       & 0.50 & 0.39 & 0.72 & \yellowc 0.75 & \orangec 0.42 & 0.52 & \orangec 0.83 & \yellowc 0.43 & \yellowc 0.60 & 0.61 & 0.74 & \yellowc 0.51 & \yellowc 0.57 \\
AltPO     & 0.66 & \yellowc 0.93 & \yellowc 0.95 & \redc 0.78 & \orangec 0.42 & \orangec 0.54 & 0.80 & \yellowc 0.43 & \orangec 0.61 & 0.61 & 0.75 & 0.47 & \yellowc 0.57 \\
GradDiff  & 0.42 & 0.35 & 0.67 & 0.73 & 0.40 & \yellowc 0.53 & \redc 0.84 & 0.42 & \redc 0.62 & \redc 0.79 & \redc 0.88 & \redc 0.53 & \redc 0.59 \\
IDKDPO    & \yellowc 0.87 & \redc 1.00 & \orangec 0.96 & 0.41 & \orangec 0.42 & \yellowc 0.53 & 0.68 & \yellowc 0.43 & 0.58 & \yellowc 0.62 & 0.75 & 0.50 & 0.52 \\
IDKNLL    & \redc 0.98 & 0.45 & 0.74 & 0.70 & 0.39 & 0.49 & 0.73 & \yellowc 0.43 & 0.57 & \orangec 0.66 & \orangec 0.78 & 0.49 & 0.55 \\
UNDIAL    & 0.69 & 0.82 & \orangec 0.96 & 0.50 & 0.38 & 0.48 & 0.78 & 0.41 & 0.56 & 0.56 & 0.61 & 0.46 & 0.51 \\
NPO       & 0.61 & 0.71 & 0.91 & \orangec 0.76 & \yellowc 0.41 & 0.52 & 0.79 & \yellowc 0.43 & \yellowc 0.60 & \orangec 0.66 & \orangec 0.78 & 0.50 & \yellowc 0.57 \\
SimNPO    & 0.65 & \orangec 0.94 & 0.94 & \redc 0.78 & \orangec 0.42 & \yellowc 0.53 & \orangec 0.83 & \redc 0.45 & \orangec 0.61 & 0.56 & 0.71 & 0.48 & 0.56 \\
EvoMU     & \orangec 0.96 & \redc 1.00 & \redc 0.97 & 0.60 & \redc 0.44 & \redc 0.58 & \yellowc 0.81 & \orangec 0.44 & \redc 0.62 & 0.60 & \yellowc 0.77 & \orangec 0.52 & \orangec 0.58  \\ \hline
\ours{} (TOFU-5\%)  & 0.77 & 1.00 & 0.92 & 0.60 & 0.43 & 0.56 & 0.80 & 0.43 & 0.63 & 0.67 & 0.81 & 0.52 & 0.58 \\
\bottomrule
\end{tabular}
\end{adjustbox}
\label{tab:tofu10}
\end{table*}

\subsection{Baselines}

In this section we briefly describe the algorithms we compare \ours{} against.

\begin{itemize}

    \item \textbf{GA}~\citep{ga, maini2024tofu}: Gradient Ascent fine-tunes the model to decrease likelihood on the forget set, pushing the model away from forbidden completions.

    \item \textbf{RMU}~\citep{li2024wmdp}: Representation Misdirection for Unlearning steers latent representations of forget tokens toward a fixed random vector while regularizing retain tokens to match the original model’s representations.

    \item \textbf{RKLD}~\citep{rkld}: Reverse KL-Divergence-based Knowledge Distillation performs unlearning by training a student model with a reverse-KL loss on forget queries so that it imitates a privacy-sanitized teacher distribution.

    \item \textbf{AltPO}~\citep{mekala2025altpo}: Alternate Preference Optimization applies a DPO-style preference loss over pairs of original and alternate answers on forget prompts .

    \item \textbf{GradDiff}~\citep{yao2024large, maini2024tofu, liu2022continual}: Gradient Difference minimizes a loss equal to the retain loss minus a scaled forget loss so that gradient descent corresponds to ascent on forget samples and descent on retain samples.
    
    \item \textbf{TaskVector}~\citep{ilharco2022editing}: Task arithmetic constructs a task vector as the parameter difference between a fine-tuned model and the base model, and applies or subtracts this vector to edit behavior.

    \item \textbf{IDKDPO / IDKNLL}~\citep{maini2024tofu}: Train the model to answer forget prompts with ``I don't know'', using a DPO-style preference loss (IDKDPO) or a standard NLL loss on the IDK target (IDKNLL).

    \item \textbf{UNDIAL}~\citep{undial}: Performs self-distillation with adjusted logits by subtracting a one-hot vector for target tokens from the teacher logits on forget data and training a student model to match these modified logits.

    \item \textbf{NPO}~\citep{npo}: Negative Preference Optimization uses a DPO-like negative-preference loss that contrasting log-probabilities of undesirable forget responses under the unlearned and reference models. 

    \item \textbf{SimNPO}~\citep{simnpo}: SimNPO simplifies NPO by removing explicit dependence on a reference model and reweighting gradients in a simpler preference-optimization loss applied to forget data.
\end{itemize}

\subsection{Results}
\paragraph{TOFU.}
TOFU (Task of Fictitious Unlearning for LLMs)~\citep{maini2024tofu} is a biographical question--answering benchmark designed to test selective forgetting.
TOFU consists of a forget split containing prompts whose underlying biographical facts should be removed, and a retain split containing similar fictitious-author questions that should remain answerable.
The TOFU-$x\%$ setting specifies that we unlearn $x\%$ of the authors while preserving performance on the remaining $(1-x)\%$.
Additionally, TOFU evaluates collateral damage using two out-of-distribution general-knowledge subsets, Real Authors and World Facts.

We report unlearning efficacy on the forget split using ROUGE-L~\citep{lin2004rouge}, Prob, and extraction strength~\citep{carlini2021extracting}, and utility preservation on the retain split and the two general-knowledge subsets.
Prob denotes the average teacher-forced probability assigned to the ground-truth answer. We report $1-\textit{Prob}$ on the forget split and Prob on the utility splits.
Following prior work, we summarize the overall forgetting-utility trade-off using the TOFU model-utility (MU) score.
We evaluate two settings, TOFU-5\% and TOFU-10\%, following~\citep{simnpo}.

As shown in \cref{tab:tofu5}, on TOFU-5\% \ours{} achieves near-saturated forgetting 
while maintaining strong utility on retain, Real Authors, and World Facts, yielding the best MU among the compared methods.
In \cref{tab:tofu10} we see that TOFU-10\% is substantially more challenging: several baselines over-unlearn and incur large utility drops (e.g., IDKDPO and UNDIAL), while others preserve utility but under-forget (e.g., RMU), while \ours{} provides highest MU. 
Additional experimental details are provided in Appendix~\ref{app:tofu}.
\paragraph{MUSE.}
MUSE \citep{shi2024muse} targets long-form unlearning where models may leak verbatim passages and underlying facts. MUSE includes two corpora (News and Books) with qualitatively different forgetting scenarios.

MUSE News consists of BBC news articles that are split into disjoint forget, retain, and holdout sets.
This setting evaluates whether an unlearning method can remove specific articles at scale while preserving performance on other news articles.
MUSE Books uses the Harry Potter book series as the forget set, while the retain set is constructed from Harry Potter FanWiki pages. This setting evaluates a harder, entangled scenario: the model should stop reproducing copyrighted book text and facts learned from it, but should still answer questions about closely related in-domain material that remains permissible.

We report three unlearning metrics on $\mathcal{D}_{\mathrm{forget}}$: VerbMem, KnowMem, and PrivLeak. We use KnowMem on $\mathcal{D}_{\mathrm{retain}}$ as the utility measure. VerbMem measures verbatim regurgitation of the forget text, KnowMem measures whether the model still reproduces the knowledge contained in the forget corpus, and PrivLeak measures membership leakage using the holdout set as a non-member reference distribution (holdout is separate from the retain set). Additional metric details are provided in Appendix~\ref{app:muse}.

Results are shown in \cref{tab:muse-unlearning}. On MUSE News, \ours{} substantially reduces privacy leakage 
while matching the best retain-set utility 
On MUSE Books, several baselines achieve low memorization on the forget set but severely degrade retain utility, whereas \ours{} preserves much higher retain performance 
while reducing memorization of the forget corpus. 

\paragraph{WMDP.}
WMDP \citep{wmdp} focuses on suppressing hazardous knowledge. 
Forgetting is measured as a drop in performance on a safety-critical question set. 

We follow the evaluation protocol of \citep{simnpo}. As the forgetting metric, we measure accuracy on WMDP-Bio and report $1-\mathrm{AccBio}$, where larger values indicate stronger suppression of the hazardous slice. As the utility metric we report accuracy on MMLU \citep{hendrycks2020measuring} to assess whether broad general knowledge is retained. Results are shown in \Cref{tab:wmdp}. 
Several baselines achieve substantial suppression with degraded utility. 
\ours{} improves upon NPO in utility 
while matching or improving forgetting.
\ours{} substantially improves over the best baseline SimNPO in utility
at the same level of forgetting. 

\begin{figure*}[ht]
\begin{subfigure}[t]{0.65\textwidth}
  \begin{minipage}[t]{0.49\textwidth}
    \centering
    \includegraphics[width=\linewidth]{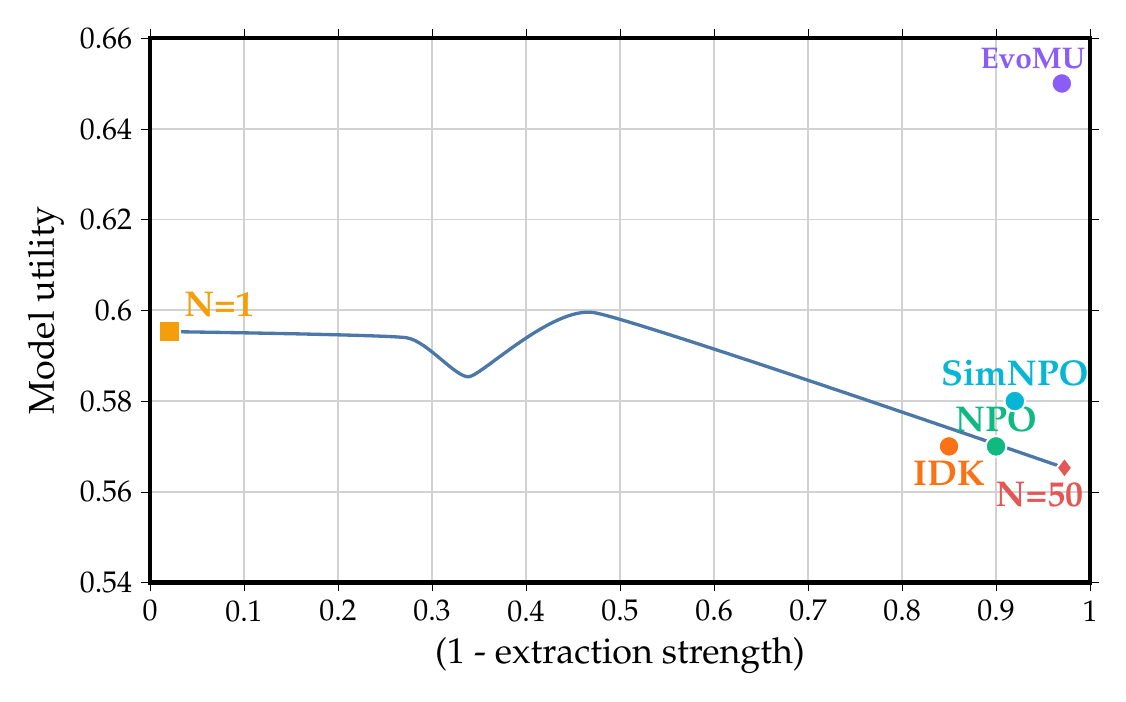}
  \end{minipage}
  \begin{minipage}[t]{0.49\textwidth}
    \centering
    \includegraphics[width=\linewidth]{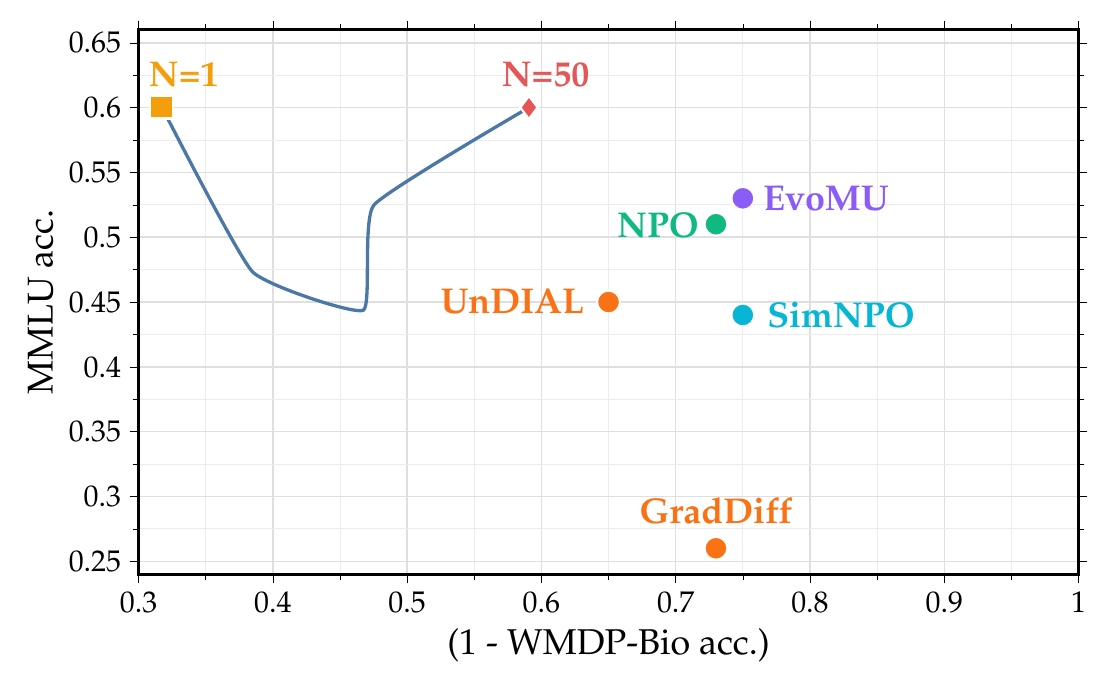}
  \end{minipage}
\caption{\underline{Left}: TOFU-5\% random-loss sampling ablation and \underline{Right}: WMDP random-loss sampling ablation. In both plots, points denote baseline unlearning methods. The curve connects the best-performing randomly sampled losses from the LLM Proposer as the number of sampled loss functions $N$ increases (shown for $N\in\{1,10,15,25,50\}$, with endpoints annotated). 
Better performance lies in the top-right region (higher utility and stronger forgetting).}
  \label{fig:plot-and-wmdp}
\end{subfigure}
\hfill
\begin{subfigure}[t]{0.32\textwidth}
\centering
  \includegraphics[width=0.95\linewidth]{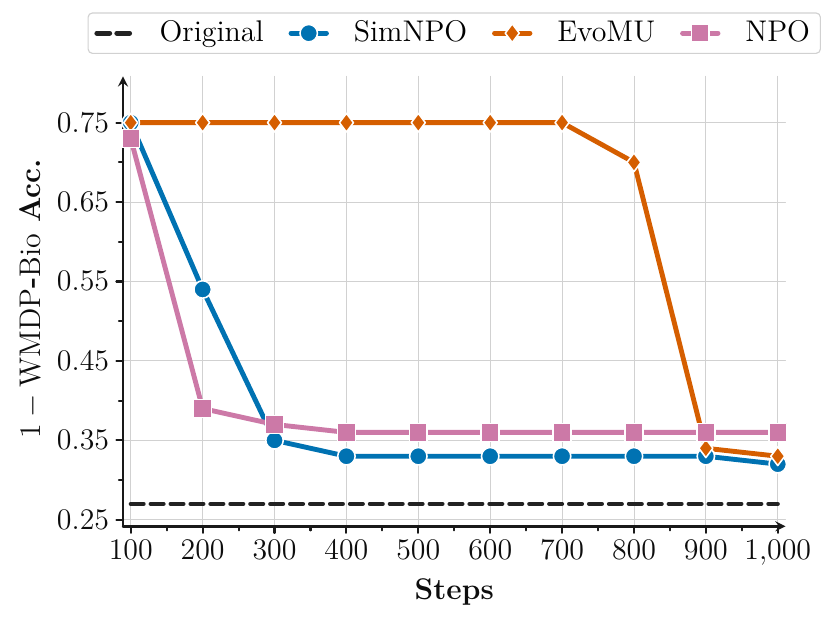}
\caption{Relearning on WMDP. We compare \ours{} to SimNPO and NPO by fine-tuning each unlearned model. The y-axis reports forgetting as $1-\mathrm{Acc}_{\text{WMDP-Bio}}$}
  \label{fig:relearning}
\end{subfigure}
\end{figure*}

\begin{table}[ht]
\centering
\setlength{\tabcolsep}{3pt}
\renewcommand{\arraystretch}{1.1}
\caption{Performance of unlearning methods on MUSE News (LLaMA2-7B) and MUSE Books (ICLM-7B). 
}
\label{tab:muse-unlearning}

\begin{adjustbox}{max width=0.9\columnwidth}
\begin{tabular}{lccc|c}
\toprule
& \multicolumn{3}{c}{Unlearning Efficacy} & Utility \\
\cmidrule(r){2-4}\cmidrule(l){5-5}
Method
& \makecell[c]{VerbMem\\$D_f$ ($\downarrow$)}
& \makecell[c]{KnowMem\\$D_f$ ($\downarrow$)}
& \makecell[c]{PrivLeak\\($\to 0$)}
& \makecell[c]{KnowMem\\$D_r$ ($\uparrow$)} \\
\midrule
\multicolumn{5}{c}{\textbf{MUSE News}} \\ \midrule

Original          &  58.29 &  62.93 & -98.71 & 54.31 \\

Retain          &  20.75 &  33.32& 0.00 & 53.79 \\

\midrule

GA          & \redc 0.00 & \redc 0.00 & \orangec 20.14 & 0.00 \\
GradDiff    & 4.85 & \orangec 31.29 & 108.12 & 28.21 \\
Task Vector & 77.42 & 58.76 & -100.00 & \orangec 47.94 \\
NPO         & \yellowc 2.53 & 56.93 & 108.91 & 37.58 \\
SimNPO      & \orangec 2.34 & \yellowc 44.84 & \yellowc 72.93 & \yellowc 39.65 \\
\ours{}     & 18.56 & 51.24 & \redc 4.38 & \redc 47.98 \\ \hline
\ours{} (TOFU-5\%)    & 42.24 & 64.19 & -99.78 & 50.51 \\


\midrule
\multicolumn{5}{c}{\textbf{MUSE Books}} \\
\midrule

Original          & 99.56 &  58.32 & -56.32 & 67.01 \\

Retain          &  14.30 &  28.90 & 0.00 & 74.50 \\ 
\midrule
GA          & \redc 0.00 & \redc 0.00 & \orangec -24.07 & 0.00 \\
GradDiff    & \redc 0.00 & \redc 0.00 & \yellowc -24.59 & 0.13 \\
Task Vector & 99.31 & \yellowc 35.55 & -83.78 & \redc 62.55 \\
NPO         & \redc 0.00 & \redc 0.00 & -31.17 & 23.71 \\
SimNPO      & \redc 0.00 & \redc 0.00 & \redc -19.82 & \yellowc 48.27 \\
\ours{}     & \redc 0.00 & \orangec 7.33 & -27.34 & \orangec 53.58 \\ \hline
\ours{} (TOFU-5\%)    & 40.01 & 28.22 & -52.68 & 46.83 \\
\bottomrule
\end{tabular}
\end{adjustbox}
\end{table}

All proposed loss functions can be found in Appendix \ref{sec:losses}.
We also include discussions on why we found those losses relevant per each dataset in Appendix \ref{app:discussion}.

\paragraph{Hyperparameter search of existing methods.}
To show that \ours{} does not merely tune existing baselines but synthesizes new loss functions, we take the strongest existing loss (SimNPO) and fit hyperparameters via an extensive grid search on TOFU-10\%.
Results are reported in Appendix~\ref{app:hyperparams}.

Across the searched configurations, SimNPO spans a range of forgetting-utility trade-offs, confirming that careful tuning can shift the operating point.
However, even the best tuned SimNPO runs remain substantially below \ours{} in overall performance. 
This suggests that the gains of \ours{} are not primarily attributable to better hyperparameters of an existing loss, but to the loss structure itself.

\paragraph{Relearning.}
Recent work suggests that many unlearning procedures are reversible: fine-tuning on a small amount of the forgotten data can rapidly restore suppressed capabilities or memorized content \citep{hu2025jogging, xu2025reversibility, xu2025relearn, fan2025towards, wang2025invariance}. Following the relearning protocol of \citep{simnpo}, we test how quickly hidden WMDP-Bio knowledge returns after unlearning. Concretely, starting from the unlearned checkpoints produced by SimNPO, NPO, and \ours{}, we sample 20\% of the forget set and fine-tune each model for 1000 steps on this subset with learning rate $10^{-5}$. We report the forgetting score $1-\mathrm{Acc}_{\text{WMDP-Bio}}$ throughout training in \cref{fig:relearning}. Consistent with prior observations \citep{simnpo}, both SimNPO and NPO lose forgetting rapidly as fine-tuning proceeds.
In contrast, \ours{} remains nearly flat for roughly the first 800 steps, suggesting substantially stronger resistance to relearning. 
After sufficient updates, all methods eventually converge to similar levels.
This experiment indicates that \ours{} forgets unlearned data more thoroughly.
 

\subsection{Ablations}     

\paragraph{Generalization of loss functions.}
We might suspect that \ours{} discovers benchmark-specific losses and may therefore overfit to a particular evaluation protocol.
To probe cross-benchmark generalization, we take the loss discovered on TOFU-5\% and apply it unchanged to the other settings: TOFU-10\%, MUSE News, MUSE Books, and WMDP.
We report the resulting performance in Tables~\ref{tab:tofu10}, \ref{tab:muse-unlearning}, and \ref{tab:wmdp} under the label \ours{} (TOFU-5\%).
We do not modify hyperparameters nor functional form of the TOFU-discovered loss. Overall, the TOFU-5\% loss transfers well to TOFU-10\% and WMDP, which argues against simple benchmark overfitting.
In contrast, performance degrades on MUSE News and MUSE Books.
We attribute this gap to the different nature of MUSE (long-form memorization, membership leakage) that makes different losses necessary.

\paragraph{Without Evolutionary Search.}
To assess importance of the evolutionary search, we compare against sampling only the initial $N$ candidates from the LLM Proposer for TOFU-5\% and WMDP. In \Cref{fig:plot-and-wmdp} we plot best results achieved with different $N$ (from $N{=}1$ to $N{=}50$). Overall MU is significantly better after evolutionary search. Interestingly, random sampling seems to already match previous SotA of our baselines on TOFU-5\%.

Interestingly, some sampled losses are not sensible, for example, they increase the log-likelihood of the forget targets, in contradiction with the unlearning goal. We provide representative examples in Appendix~\ref{app:nonsense}. Notably, such pathological losses do not appear among the final-generation candidates: the evolutionary selection process consistently filters them out.

Additional ablation experiments ($\#$ thinking tokens, impact of proposing $\#$ of epochs to train, robustness w.r.t.\ seeds) can be found in Appendix~\ref{sec:app_ablations}.

\begin{table}
\caption{Results on the WMDP (Llama-3-8B-Instruct). 
Forgetting is measured by $1-\mathrm{AccBio}$ on WMDP-Bio and utility by accuracy on MMLU.
}
\centering
\setlength{\tabcolsep}{4pt}
\renewcommand{\arraystretch}{1.05}
\resizebox{\columnwidth}{!}{%
\begin{tabular}{lcc}
\toprule
& \multicolumn{1}{c}{Unlearning Efficacy} & \multicolumn{1}{c}{Utility Preservation} \\
\cmidrule(lr){2-2}\cmidrule(lr){3-3}
Method & $1-\mathrm{AccBio}\,(\uparrow)$ & $\mathrm{MMLU}\,(\uparrow)$ \\
\midrule
Original & 0.27 & 0.65 \\ \hline
UnDIAL   & 0.65 & 0.45 \\
GradDiff & \orangec 0.73 & 0.26 \\
IDKNLL   & 0.66 & \yellowc 0.47 \\
IDKDPO   & \yellowc 0.67 & 0.44 \\
NPO      & \orangec 0.73 & \orangec 0.51 \\
SimNPO   & \redc 0.75 & 0.44 \\
\ours{}  & \redc 0.75 & \redc 0.53 \\ \hline
\ours{} (TOFU-5\%) & 0.67 & 0.62 \\
\bottomrule
\end{tabular}%
}
\label{tab:wmdp}
\end{table}

\section{Conclusions \& Limitations}
We have proposed an evolutionary search procedure that automatically synthesizes dataset-specific unlearning losses using LLMs as co-scientists.
We use comparatively little compute to come up with losses, yet outperform existing ones proposed by deep learning experts.
More broadly, we argue that it might be more advantageous to synthesize specific components of deep learning systems that can be evaluated than to design them by hand.
We conjecture that in many applications similar to machine unlearning the design space might be large enough so that a search procedure such as ours will be able to discover new and superior solutions.

\clearpage

\section*{Impact Statement}
Our work \ours{} makes machine unlearning more practical by automatically discovering \emph{task-specific} fine-tuning losses that better balance targeted forgetting with retained capability, reducing the need for brittle hand-tuning and potentially lowering the cost of responding to data-deletion requests or mitigating unwanted memorization (e.g., sensitive, harmful, or copyrighted content) without full retraining; additionally, because the discovered loss code and evaluation results are explicit artifacts, the process can be more auditable and reproducible than ad hoc loss design. At the same time, responsible use still requires careful evaluation and access controls, since unlearning methods can be dual-use (e.g., repurposed to remove safety behaviors) and benchmark-driven optimization can give misleading confidence if metrics do not capture real deployment threat models, so we view \ours{} as a tool that strengthens the unlearning workflow when paired with rigorous, transparent evaluation.

\section*{Acknowledgments}
The authors gratefully acknowledge the funding of this project by computing time provided by the Paderborn Center for Parallel Computing (PC2).

\newpage
\appendix
\onecolumn

\section{Pseudocode}
\label{app:pseudocode}

\begin{algorithm}[ht]
\caption{\ours{}}
\label{alg:search}
\begin{algorithmic}[1]
\REQUIRE Base model $p_{\theta_0}$; forget/retain datasets $\mathcal{D}_f,\mathcal{D}_r$; generator $\mathcal{G}$;
population size $N$; iterations $R$; top-$K$ parents; children per parent $C$
\STATE \textbf{(Optional)} Precompute reference caches $\mathbf{z}^{\mathrm{ref}}_f,\mathbf{z}^{\mathrm{ref}}_r$ using $p_{\theta_0}$
\STATE $\mathcal{S}_0 \leftarrow \mathrm{Generate}(\{\mathcal{G}(\emptyset)\}_{i=1}^{N}$ \hfill (initial losses)
\STATE $\mathcal{B} \leftarrow \emptyset$ \hfill (best-so-far record)
\FOR{$t = 0,\dots,R-1$}
  \FOR{each candidate loss $\mathcal{L} \in \mathcal{S}_t$}
    \STATE Parse training budget (e.g., epochs/steps) from $\mathcal{L}$'s metadata
    \STATE Train LoRA adapters from $p_{\theta_0}$ on $\mathcal{D}_f,\mathcal{D}_r$ minimizing $\mathcal{L}$
    \STATE Merge LoRA weights into the base model to obtain checkpoint $p_{\theta_{\mathcal{L}}}$
    \STATE Evaluate $p_{\theta_{\mathcal{L}}}$ with the unlearning benchmark; extract metrics $m(\mathcal{L})$ and training history $H(\mathcal{L})$
    \STATE Compute selection score $s(\mathcal{L}) \leftarrow \mathrm{Score}(m(\mathcal{L}))$
    \STATE Update best-so-far $\mathcal{B} \leftarrow \arg\max_{\mathcal{L}' \in \mathcal{B}\cup\{\mathcal{L}\}} s(\mathcal{L}')$
  \ENDFOR
  \STATE $\mathcal{P}_t \leftarrow \mathrm{TopK}(\mathcal{S}_t; s)$ \hfill (parent selection)
  \STATE $\mathcal{S}_{t+1} \leftarrow \mathrm{Refine}\Big(\!\bigcup_{\mathcal{L}\in\mathcal{P}_t} \{\mathcal{G}(\mathcal{L}, m(\mathcal{L}), H(\mathcal{L}))\}_{j=1}^{C}\Big)$ \hfill (refine top-K selection)
\ENDFOR
\STATE \textbf{return} best loss $\mathcal{L}^\star$ and its checkpoint $p_{\theta_{\mathcal{L}^\star}}$ from $\mathcal{B}$
\end{algorithmic}
\end{algorithm}

\section{Obtained loss functions}
\label{sec:losses}

In this section we present the loss functions obtained for each benchmark. 

\begin{promptbox}{Gray}{TOFU5\% (7 epochs)}
$\ell(\theta)
=
\beta\,\mathbb{E}_{(x_f,y_f)\sim D_f}\!\Big[
\log \pi_\theta(y_f\!\mid x_f) - \log \pi_{\mathrm{ref}}(y_f\!\mid x_f)
\Big]+
\mathbb{E}_{(x_r,y_r)\sim D_r}\!\Big[
\log \pi_{\mathrm{ref}}(y_r\!\mid x_r) - \log \pi_\theta(y_r\!\mid x_r)
\Big],
\qquad \beta=1.2.$ 

\end{promptbox}

\begin{promptbox}{Gray}{TOFU10\% (8 epochs)}
$\ell(\theta)
=
\mathbb{E}_{(x_f,y_f)\sim D_f,\,(x_r,y_r)\sim D_r}\Big[
\exp\!\Big(\log \pi_\theta(y_f\!\mid x_f)-\log \pi_{\mathrm{ref}}(y_f\!\mid x_f)\Big)
- 2\rho\,
\exp\!\Big(\log \pi_\theta(y_r\!\mid x_r)-\log \pi_{\mathrm{ref}}(y_r\!\mid x_r)\Big)
\Big],
\qquad \rho=0.3.$

\end{promptbox}

\begin{promptbox}{Gray}{MUSE-Books (8 epochs)}
$\ell(\theta)
=
\mathbb{E}_{(x_f,y_f)\sim D_f,\,(x_r,y_r)\sim D_r}\Big[
0.7\,\log \pi_\theta(y_f\!\mid x_f)
-\log \pi_\theta(y_r\!\mid x_r)
+\alpha\,\max\!\Big(\log \pi_{\mathrm{ref}}(y_r\!\mid x_r)-\log \pi_\theta(y_r\!\mid x_r),\,0\Big)
\Big],
\qquad \alpha=0.3.$

\end{promptbox}

\begin{promptbox}{Gray}{MUSE-News (8 epochs)}
$\ell(\theta)
=
0.35\,
\mathbb{E}_{(x_f,y_f)\sim D_f,\,(x_r,y_r)\sim D_r}\Big[
\min\!\Big(
\log \pi_\theta(y_f\!\mid x_f)-\log \pi_\theta(y_r\!\mid x_r),
\,1
\Big)
\Big].$
\end{promptbox}

\begin{promptbox}{Gray}{WMDP (300 steps)}
$\ell(\theta)
=
\mathbb{E}_{(x_f,y_f)\sim D_f,\,(x_r,y_r)\sim D_r}\Big[
1.5\big(\log \pi_\theta(y_f\!\mid x_f)-\log \pi_{\mathrm{ref}}(y_f\!\mid x_f)\big)
-\big(\log \pi_\theta(y_r\!\mid x_r)-\log \pi_{\mathrm{ref}}(y_r\!\mid x_r)\big)
\Big].$
\end{promptbox}

\section{Proposed Loss Functions Discussion}
\label{app:discussion}

This section explains why the discovered losses in Appendix~\ref{sec:losses} are well matched to each benchmark and why they can yield better forgetting--utility trade-offs than NPO/SimNPO-style losses. A common pattern across all discovered objectives is that they (i) impose explicit pressure to reduce the model’s likelihood on the forget targets, and (ii) simultaneously include an explicit retention term that either increases likelihood on retain targets or prevents retain likelihood from falling below a reference. This explicit two-sided control is important because most failures in unlearning are not due to insufficient forgetting, but due to collateral utility loss from overly aggressive updates.

\paragraph{TOFU-5\% objective.}

Proposed loss function: 
$$
\ell(\theta)
=
\beta\,\mathbb{E}_{(x_f,y_f)\sim D_f}\!\Big[
\log \pi_\theta(y_f\!\mid x_f) - \log \pi_{\mathrm{ref}}(y_f\!\mid x_f)
\Big]+
\mathbb{E}_{(x_r,y_r)\sim D_r}\!\Big[
\log \pi_{\mathrm{ref}}(y_r\!\mid x_r) - \log \pi_\theta(y_r\!\mid x_r)
\Big],
\qquad \beta=1.2. 
$$

The TOFU-5\% loss is a reference-anchored, sign-asymmetric delta objective:
it penalizes the model when it assigns higher log-probability than the reference to the forget answers, while rewarding the model for assigning higher log-probability than the reference to the retain answers. Concretely, the forget term has the form
$\log \pi_\theta(y_f\mid x_f)-\log \pi_{\mathrm{ref}}(y_f\mid x_f)$ with a positive weight, so minimizing the loss drives this delta negative, i.e., pushes $\pi_\theta(y_f\mid x_f)$ below the reference.
This directly targets the teacher-forced probability metric used by TOFU and, by reducing the model’s likelihood of the ground-truth answer, also reduces extraction strength, which is computed from length-normalized likelihood under extraction prompts.

The retain term has the opposite sign,
$\log \pi_{\mathrm{ref}}(y_r\mid x_r)-\log \pi_\theta(y_r\mid x_r)$,
so minimizing it pushes the retain delta positive (increase retain likelihood above the reference).
This is a useful inductive bias for TOFU-5\%, where the forget set is relatively small and the primary risk is over-unlearning: the objective encourages the model to compensate for the forgetting pressure by explicitly improving retain likelihood rather than merely trying to stay close in KL.
The single scalar $\beta$ sets the forgetting--utility balance without requiring paired preferences or synthetic negative answers.

Compared to NPO/SimNPO, this objective provides tighter control over retention. NPO/SimNPO-style losses primarily emphasize lowering undesired likelihood on forget responses; retention is typically handled indirectly (e.g., via auxiliary retain NLL, regularization, or reference comparisons that are symmetric in form). The discovered TOFU-5\% objective is intentionally asymmetric: it uses the reference model as a stable anchor for forget suppression, but it actively pushes retain likelihood upward, which helps preserve (and sometimes improve) the MU components based on retain/Real-Authors/World-Facts probabilities.

\paragraph{TOFU-10\% objective.}

Proposed loss function:

$$
\ell(\theta)
=
\mathbb{E}_{(x_f,y_f)\sim D_f,\,(x_r,y_r)\sim D_r}\Big[
\exp\!\Big(\log \pi_\theta(y_f\!\mid x_f)-\log \pi_{\mathrm{ref}}(y_f\!\mid x_f)\Big)
- 2 \cdot 0.3\,
\exp\!\Big(\log \pi_\theta(y_r\!\mid x_r)-\log \pi_{\mathrm{ref}}(y_r\!\mid x_r)\Big)
\Big].
$$

TOFU-10\% is harder because many more entities are removed, so simple linear penalties can under-focus on the remaining hard-to-forget examples or can require large coefficients that increase collateral damage. The discovered objective replaces linear deltas with exponentiated deltas:
$\exp(\log\pi_\theta-\log\pi_{\mathrm{ref}})$ on both forget and retain, with opposite signs.
Minimization drives the forget delta negative (reducing $\exp(\Delta_f)$) and encourages the retain delta positive (increasing $\exp(\Delta_r)$ because it is subtracted).

The exponential transform acts as an adaptive reweighting mechanism: samples with positive deltas (cases where the model is still more confident than the reference) receive disproportionately larger gradients. On TOFU-10\%, this concentrates optimization on stubborn memorized facts without requiring manual per-sample weighting or brittle margins. At the same time, the retain exponential term creates strong counter-pressure that discourages global degradation of likelihood on the retain distribution, which is exactly the failure mode observed in many over-unlearning baselines at higher deletion rates.

Relative to NPO/SimNPO, the key difference is that the objective shapes gradient allocation across examples through the exponential rather than through preference comparisons. NPO/SimNPO is effective when the preference signal (undesirable vs. alternate) is well specified, but in TOFU the evaluation is teacher-forced likelihood and extraction prompts; exponentiated deltas provide a direct, automatically scaled mechanism aligned with these metrics while maintaining a strong, explicit retain reward.

\paragraph{MUSE News objective.}

Proposed loss function:

$$
\ell(\theta)
=
0.35\,
\mathbb{E}_{(x_f,y_f)\sim D_f,\,(x_r,y_r)\sim D_r}\Big[
\min\!\Big(
\log \pi_\theta(y_f\!\mid x_f)-\log \pi_\theta(y_r\!\mid x_r),
\,1
\Big)
\Big].
$$

MUSE News is a deletion-at-scale scenario within a relatively coherent domain distribution: the forget and retain documents are similar in style and topic (same source distribution), and the benchmark additionally evaluates membership leakage using a separate holdout set. In this setting, a good objective should avoid large distribution shifts while still making the forget documents behave like non-members.

The discovered objective is a capped relative-ranking loss based on the difference between forget and retain log-probabilities:
$\min(\log\pi_\theta(y_f\mid x_f)-\log\pi_\theta(y_r\mid x_r),\,1)$ (up to a scaling constant).
Minimizing encourages $\log\pi_\theta(y_f\mid x_f)$ to be smaller than $\log\pi_\theta(y_r\mid x_r)$, i.e., it pushes the model to treat forget targets as less likely than retain targets, rather than enforcing an absolute decrease relative to a reference.
This relative formulation is well matched to MUSE News because it implicitly normalizes for domain-wide calibration shifts: even if the model’s overall likelihood scale changes slightly during fine-tuning, the loss keeps forget likelihood suppressed relative to retain likelihood, which helps reduce membership signals without collapsing general in-domain competence.

The cap (the $\min(\cdot,1)$) prevents extreme gradients when $\log\pi(y_f)-\log\pi(y_r)$ becomes very large, which improves stability and reduces the chance of runaway updates that damage utility.
This is a concrete way to mitigate over-unlearning in a setting where the model is trained on long-form data and where large updates can easily cause broad degradation.

Compared to NPO/SimNPO, this objective does not require a reference model or preference pairs and instead uses the retain batch as an implicit anchor. In MUSE News, that anchor is particularly appropriate because retain is drawn from the same distribution as forget. NPO/SimNPO can drift (SimNPO) or overly constrain to the reference (NPO) in ways that do not directly target membership leakage; the discovered ranking-and-cap structure more directly targets the relative separability that underlies membership inference while preserving in-domain behavior.

\paragraph{MUSE Books objective.}

Proposed loss function:

$$
\ell(\theta)
=
\mathbb{E}_{(x_f,y_f)\sim D_f,\,(x_r,y_r)\sim D_r}\Big[
0.7\,\log \pi_\theta(y_f\!\mid x_f)
-\log \pi_\theta(y_r\!\mid x_r)
+0.3\,\max\!\Big(\log \pi_{\mathrm{ref}}(y_r\!\mid x_r)-\log \pi_\theta(y_r\!\mid x_r),\,0\Big)
\Big]
$$

MUSE Books is qualitatively different from MUSE News: the forget set is copyrighted book text, while the retain set is constructed from closely related material (FanWiki pages) that overlaps heavily in entities, relationships, and surface cues. This creates an entangled unlearning problem: the model must stop reproducing verbatim passages and book-derived facts, but still answer questions about the same universe using permissible retain sources. In such a setting, objectives that simply push down likelihood on forget targets can easily harm retain performance because the representations and lexical triggers are shared.

The discovered Books objective combines three ingredients:
(1) a weighted forget penalty ($0.7\,\log\pi_\theta(y_f\mid x_f)$) that reduces book memorization pressure without being excessively aggressive,
(2) a strong retain reward ($-\log\pi_\theta(y_r\mid x_r)$) that explicitly increases likelihood on the permissible in-domain retain corpus, and
(3) a one-sided hinge barrier on retain relative to the reference,
$\alpha \max(\log\pi_{\mathrm{ref}}(y_r\mid x_r)-\log\pi_\theta(y_r\mid x_r),0)$.
The hinge activates only when retain likelihood drops below the reference, acting like a safety constraint that prevents the model from sacrificing retain knowledge to achieve lower forget likelihood.

This structure makes sense for Books because the main risk is not insufficient forgetting, but damaging closely related retain competence. The hinge term is a targeted guardrail: it does not force the model to stay identical to the reference on retain, but it prevents regressions. That asymmetry is valuable in an overlapping-domain scenario because it allows the optimizer to reshape behavior where needed (reduce book memorization) while preserving or improving behavior on the permissible corpus.

Relative to NPO/SimNPO, the hinge barrier provides a more direct mechanism for retain protection. Preference-style losses can be effective for discouraging particular outputs, but when forget and retain are highly entangled, suppressing the undesired outputs often suppresses the desired ones as well unless the objective includes an explicit constraint. The discovered loss encodes such a constraint directly at the level of retain likelihood, which is closely aligned with the KnowMem utility metric on $\mathcal{D}_{\mathrm{retain}}$.

\paragraph{WMDP objective.}

Proposed loss function:

$$
\ell(\theta)
=
\mathbb{E}_{(x_f,y_f)\sim D_f,\,(x_r,y_r)\sim D_r}\Big[
1.5\big(\log \pi_\theta(y_f\!\mid x_f)-\log \pi_{\mathrm{ref}}(y_f\!\mid x_f)\big)
-\big(\log \pi_\theta(y_r\!\mid x_r)-\log \pi_{\mathrm{ref}}(y_r\!\mid x_r)\big)
\Big].
$$

WMDP measures forgetting as reduced accuracy on a hazardous knowledge slice (WMDP-Bio), with utility measured on a broad capability benchmark (MMLU). Unlike TOFU or MUSE, the target is not primarily verbatim memorization; it is a transferable capability. This setting benefits from objectives that are (i) explicitly anchored to a reference model for stability, and (ii) able to apply stronger pressure on the forget slice without causing widespread degradation.

The discovered WMDP objective is a weighted reference-delta difference:
$1.5(\log\pi_\theta-\log\pi_{\mathrm{ref}})$ on forget minus $(\log\pi_\theta-\log\pi_{\mathrm{ref}})$ on retain.
Minimizing drives the forget delta negative (reduce hazardous likelihood relative to the original model) while driving the retain delta positive (preserve or improve likelihood on retained targets relative to the original model).
The heavier forget weight reflects the fact that meaningful suppression of hazardous accuracy often requires stronger targeted pressure than biography-style forgetting, while the explicit retain delta term counterbalances this to maintain general performance.

Compared to NPO/SimNPO, this objective is closer in spirit but differs in two practical ways that matter for WMDP. First, it keeps an explicit reference anchor, which reduces drift under strong forgetting pressure; SimNPO-style objectives without a reference can move the model broadly, harming MMLU. Second, it includes an explicit retain-improvement term rather than relying on indirect regularization, which helps maintain general competence when the forget objective targets a capability that shares features with general knowledge.

\paragraph{Summary of why these objectives can outperform NPO/SimNPO.}
Across benchmarks, the discovered objectives share two properties that are not guaranteed in standard NPO/SimNPO formulations: (i) explicit, often asymmetric retain protection (sometimes with a hard one-sided barrier), and (ii) loss shaping that adapts to the benchmark’s failure mode (exponential reweighting for TOFU-10\%, relative ranking and gradient capping for MUSE News, hinge-based retain constraints for MUSE Books, and reference-anchored weighted deltas for WMDP). These design choices better match the evaluation metrics used in each benchmark and directly target over-unlearning, which is the dominant reason strong baselines lose MU or general utility in our experiments.

\section{Robustness-run objective definitions}
\label{app:robust_losses}

We list the exact objectives used in the five-seed robustness experiment. Each loss is written in terms of
per-example average log-probabilities on forget/retain batches, $\mathbf{z}_f,\mathbf{z}_r\in\mathbb{R}^B$,
and (optionally) reference-model values $\mathbf{z}^{\mathrm{ref}}_f,\mathbf{z}^{\mathrm{ref}}_r$.

\begin{promptbox}{Gray}{Loss 17 (7 epochs)}
$\ell_{17}(\theta)
=
\frac{1}{B}\sum_{i=1}^{B}\Big[
\max\!\big(z_f^{(i)},-10\big)
-0.4\,z_r^{(i)}
+0.6\,\max\!\big(z_f^{(i)}-z^{(i)}_{f,\mathrm{ref}},\,0\big)
\Big].$
\end{promptbox}

\begin{promptbox}{Gray}{Loss 10 (5 epochs)}
$\ell_{10}(\theta)
=
\frac{1}{B}\sum_{i=1}^{B}\Big[
z_f^{(i)}-\min\!\big(z_r^{(i)},\alpha\big)
\Big],
\qquad \alpha=0.4.$
\end{promptbox}

\begin{promptbox}{Gray}{Loss 2 (2 epochs)}
$\ell_{2}(\theta)
=
\beta\cdot \frac{1}{B}\sum_{i=1}^{B}
\Big[
\log\!\big(\exp(z_f^{(i)})+\varepsilon\big)
-\log\!\big(\exp(z^{(i)}_{f,\mathrm{ref}})+\varepsilon\big)
\Big]
+\frac{1}{B}\sum_{i=1}^{B}
\Big[
\log\!\big(\exp(z^{(i)}_{r,\mathrm{ref}})+\varepsilon\big)
-\log\!\big(\exp(z_r^{(i)})+\varepsilon\big)
\Big],
\qquad \beta=1.2,\ \varepsilon=10^{-6}.$
\end{promptbox}

\begin{promptbox}{Gray}{Loss 5 (5 epochs)}
$\ell_{5}(\theta)
=
\frac{1}{B}\sum_{i=1}^{B}\Big[
- z_r^{(i)}
+\delta\,\min\!\big(z_f^{(i)},0\big)
-0.2\,\max\!\big(z_r^{(i)}-z^{(i)}_{r,\mathrm{ref}},\,0\big)
\Big],
\qquad \delta=0.6.$
\end{promptbox}

\begin{promptbox}{Gray}{Loss 9 (3 epochs)}
$\ell_{9}(\theta)
=
\gamma\cdot \frac{1}{B}\sum_{i=1}^{B}
\Big[
\log\!\big(\exp(z_f^{(i)})+\varepsilon\big)
-\log\!\big(\exp(z^{(i)}_{f,\mathrm{ref}})+\varepsilon\big)
\Big]
+\frac{1}{B}\sum_{i=1}^{B}
\Big[
\log\!\big(\exp(z^{(i)}_{r,\mathrm{ref}})+\varepsilon\big)
-\log\!\big(\exp(z_r^{(i)})+\varepsilon\big)
\Big],
\qquad \gamma=1.5,\ \varepsilon=10^{-6}.$
\end{promptbox}

\section{Initial functions generated}
\label{app:initial_losses}

In this subsection we list the $N{=}10$ initial candidate loss functions sampled from the proposer at iteration 0.
Each objective is expressed in terms of per-example average log-probabilities on forget/retain batches,
$\mathbf{z}_f,\mathbf{z}_r\in\mathbb{R}^B$, and (optionally) reference-model values
$\mathbf{z}^{\mathrm{ref}}_f,\mathbf{z}^{\mathrm{ref}}_r$ (Eq.~\ref{eq:genericloss}). We denote the $i$-th
elements by $z_f^{(i)}$, $z_r^{(i)}$, and $z^{(i)}_{f,\mathrm{ref}}$.

\begin{promptbox}{Gray}{Initial Loss 1 (1 epoch)}
$\ell_{1}(\theta)
=
\frac{1}{B}\sum_{i=1}^{B}\Big[
- z_r^{(i)} + \alpha\, z_f^{(i)}
\Big],
\qquad \alpha=0.7.$
\end{promptbox}

\begin{promptbox}{Gray}{Initial Loss 2 (2 epochs)}
$\ell_{2}(\theta)
=
\frac{1}{B}\sum_{i=1}^{B}
\max\!\Big(
- z_r^{(i)} + \alpha\, z_f^{(i)},
\,0
\Big),
\qquad \alpha=0.5.$
\end{promptbox}

\begin{promptbox}{Gray}{Initial Loss 3 (3 epochs)}
$\ell_{3}(\theta)
=
\frac{1}{B}\sum_{i=1}^{B}\Big[
- z_r^{(i)} + \alpha\, \min\!\big(z_f^{(i)},0\big)
\Big],
\qquad \alpha=0.8.$
\end{promptbox}

\begin{promptbox}{Gray}{Initial Loss 4 (4 epochs)}
$\ell_{4}(\theta)
=
\frac{1}{B}\sum_{i=1}^{B}\Big[
- z_r^{(i)} + \alpha\, \max\!\big(z_f^{(i)}-z^{(i)}_{f,\mathrm{ref}},\,0\big)
\Big],
\qquad \alpha=0.6.$
\end{promptbox}

\begin{promptbox}{Gray}{Initial Loss 5 (5 epochs)}
$\ell_{5}(\theta)
=
\frac{1}{B}\sum_{i=1}^{B}\Big[
- z_r^{(i)} + \alpha\, \exp\!\big(z_f^{(i)}\big)
\Big],
\qquad \alpha=0.9.$
\end{promptbox}

\begin{promptbox}{Gray}{Initial Loss 6 (6 epochs)}
$\ell_{6}(\theta)
=
\frac{1}{B}\sum_{i=1}^{B}\Big[
- z_r^{(i)} + \alpha\, \sigma\!\big(z_f^{(i)}-z^{(i)}_{f,\mathrm{ref}}\big)
\Big],
\qquad \alpha=0.4,\ \ \sigma(u)=\frac{1}{1+e^{-u}}.$
\end{promptbox}

\begin{promptbox}{Gray}{Initial Loss 7 (7 epochs)}
$\ell_{7}(\theta)
=
\frac{1}{B}\sum_{i=1}^{B}\Big[
- z_r^{(i)} + \alpha\, \big|z_f^{(i)}-z^{(i)}_{f,\mathrm{ref}}\big|
\Big],
\qquad \alpha=0.3.$
\end{promptbox}

\begin{promptbox}{Gray}{Initial Loss 8 (8 epochs)}
$\ell_{8}(\theta)
=
\frac{1}{B}\sum_{i=1}^{B}\Big[
- z_r^{(i)} + \alpha\, \big(z_f^{(i)}-z^{(i)}_{f,\mathrm{ref}}\big)^{2}
\Big],
\qquad \alpha=0.2.$
\end{promptbox}

\begin{promptbox}{Gray}{Initial Loss 9 (9 epochs)}
$\ell_{9}(\theta)
=
\frac{1}{B}\sum_{i=1}^{B}\Big[
- z_r^{(i)} + \alpha\, \log\!\big(1 + (z_f^{(i)}-z^{(i)}_{f,\mathrm{ref}})\big)
\Big],
\qquad \alpha=0.1.$
\end{promptbox}

\begin{promptbox}{Gray}{Initial Loss 10 (10 epochs)}
$\ell_{10}(\theta)
=
\frac{1}{B}\sum_{i=1}^{B}\Big[
- z_r^{(i)} + \alpha\, \exp\!\big(z_f^{(i)}-z^{(i)}_{f,\mathrm{ref}}\big)
\Big],
\qquad \alpha=0.5.$
\end{promptbox}

\section{Metrics \& Model Benchmark summary}
\label{app:summary}

n this section, we present Table~\ref{tab:benchmarks}, which summarizes the base model and the evaluation metrics used for unlearning effectiveness and utility preservation in each benchmark.



\newcolumntype{Y}{>{\raggedright\arraybackslash}X}



\newcolumntype{Y}{>{\raggedright\arraybackslash}X}

\begin{table*}
\caption{Benchmarks, base models, and evaluation metrics used for unlearning effectiveness and utility preservation.}
\centering
\footnotesize
\setlength{\tabcolsep}{6pt}
\renewcommand{\arraystretch}{1.15}

\begin{adjustbox}{max width=\textwidth}
\begin{tabularx}{\textwidth}{@{} l l Y Y @{}}
\\
\midrule
\textbf{Benchmark} & \textbf{LLM to be used} &
\textbf{Unlearning Effectiveness} & \textbf{Utility Preservation} \\
\midrule

TOFU &
\makecell[l]{LLaMA-2-chat 7B} &
\begin{tabular}[t]{@{}l@{\hspace{4pt}}c@{}}
Probability on $\mathcal{D}_f$ & $\downarrow$\\
ROUGE-L on $\mathcal{D}_f$ & $\downarrow$\\
Extraction Strength on $\mathcal{D}_f$ & $\downarrow$ \\
\end{tabular}
&
\begin{tabular}[t]{@{}l@{\hspace{4pt}}c@{}}
Model utility (harmonic mean of 9 metrics) & $\uparrow$\\
Prob.\ on $\mathcal{D}_r/\mathcal{D}_{\textit{real\_author}}/\mathcal{D}_{\textit{world\_facts}}$ & $\uparrow$\\
ROUGE-L on $\mathcal{D}_r/\mathcal{D}_{\textit{real\_author}}/\mathcal{D}_{\textit{world\_facts}}$ & $\uparrow$\\
Truth ratio on $\mathcal{D}_r/\mathcal{D}_{\textit{real\_author}}/\mathcal{D}_{\textit{world\_facts}}$ & $\uparrow$
\end{tabular}
\\
\midrule

MUSE &
\makecell[l]{ICLM-7B\\LLaMA-2 7B} &
\begin{tabular}[t]{@{}l@{\hspace{4pt}}c@{}}
KnowMem on $\mathcal{D}_f$ & $\downarrow$\\
VerbMem on $\mathcal{D}_f$ & $\downarrow$\\
PrivLeak & $\to 0$
\end{tabular}
&
\begin{tabular}[t]{@{}l@{\hspace{4pt}}c@{}}
KnowMem on $\mathcal{D}_r$ & $\uparrow$
\end{tabular}
\\
\midrule

WMDP &
\makecell[l]{Llama3-8B-Instruct} &
\begin{tabular}[t]{@{}l@{\hspace{4pt}}c@{}}
Accuracy on WMDP-Bio & $\downarrow$
\end{tabular}
&
\begin{tabular}[t]{@{}l@{\hspace{4pt}}c@{}}
Accuracy on MMLU & $\uparrow$
\end{tabular}
\\

\bottomrule
\end{tabularx}
\end{adjustbox}

\label{tab:benchmarks}
\end{table*}

\section{Unlearning on TOFU}
\label{app:tofu}
In this section we showcase which unlearning metrics did we use for TOFU 5\% and 10\%. We run experiments on TOFU 5\% and 10\% with learning rate equal to 5e-4 with batch size 8. 

\paragraph{Probability.}
For each example in the retain or forget split, we compute the model’s length-normalized conditional
likelihood of the answer given the question, $P(a \mid q)^{1/|a|}$, where $q$ is the question, $a$ is the
answer, and $|a|$ is the number of tokens in $a$.
For the real authors and world facts subsets, each prompt is paired with five candidate answers
$\{a_0,\tilde{a}_1,\tilde{a}_2,\tilde{a}_3,\tilde{a}_4\}$: one correct answer $a_0$ and four perturbed (incorrect)
alternatives. We summarize preference for the correct answer via the ratio
\[
\frac{P(a_0 \mid q)^{1/|a_0|}}{\sum_{i=1}^{4} P(\tilde{a}_i \mid q)^{1/|\tilde{a}_i|}} \, .
\]

\paragraph{Truth ratio.}
The truth ratio compares the model’s preference for a paraphrased correct answer $\hat{a}$ against a set of
perturbed incorrect answers $A=\{\tilde{a}_1,\tilde{a}_2,\ldots\}$. Specifically, we take the geometric mean of
the perturbed answers’ length-normalized likelihoods and divide by the length-normalized likelihood of
$\hat{a}$:
\[
R_{\text{truth}} \;=\;
\frac{\left(\prod_{i=1}^{|A|} P(\tilde{a}_i \mid q)^{1/|\tilde{a}_i|}\right)^{1/|A|}}
{P(\hat{a} \mid q)^{1/|\hat{a}|}} \, .
\]
For the real authors and world facts subsets, no paraphrase is provided; in that case we use the
original correct answer $a$ in the denominator.

\paragraph{ROUGE-L.}
Across all TOFU subsets, we report ROUGE-L recall~\cite{lin2004rouge} between the ground-truth responses
(from the forget split) and the model outputs produced after unlearning.

\paragraph{Extraction strength.}
To quantify how easily a model can be induced to reveal information that should be forgotten, we measure
extraction strength as the model’s length-normalized likelihood of the target (ground-truth) answer
under a set of extraction-oriented prompts.
Let $E=\{e_1,\dots,e_K\}$ denote $K$ extraction templates (e.g., paraphrases or adversarially phrased requests),
and let $q^{(k)} = \mathrm{Compose}(q, e_k)$ be the resulting query for template $e_k$. We compute
\[
S_{\text{ext}}(q,a) \;=\; \max_{k \in \{1,\dots,K\}} P\!\left(a \mid q^{(k)}\right)^{1/|a|} \, ,
\]
where the maximization reflects a strong (best-of-$K$) attacker that tries multiple prompts and uses the
most effective one. Higher extraction strength indicates that the target answer is easier to extract (i.e.,
weaker unlearning), while lower values suggest better resistance to extraction.

\paragraph{Model utility.}
We aggregate overall model utility as the harmonic mean of nine quantities: the answer probability, truth
ratio, and ROUGE-L recall computed on each of the retain, real authors, and world facts
subsets. Higher utility indicates better overall performance.  

\section{Unlearning on MUSE}
\label{app:muse}

MUSE~\citep{shi2024muse} evaluates unlearning in long-form domains using memorization and privacy-leakage
metrics computed on a forget split $\mathcal{D}_{\mathrm{forget}}$ and a retain split
$\mathcal{D}_{\mathrm{retain}}$. Below we briefly summarize the metrics used in our experiments. We run experiments on MUSE News with learning rate $5e-4$ and batch size 16, and on MUSE Books with learning rate 1e-3 and batch size 4. 

\paragraph{VerbMem.}
VerbMem (verbatim memorization) measures how much the model reproduces the forget text word-for-word.
For each forget example, the model is prompted to generate a continuation/answer, and VerbMem scores the
extent of exact-string or near-exact overlap between the generated text and the reference forget span
(e.g., using longest common subsequence or high-threshold $n$-gram overlap as defined in~\citep{shi2024muse}).
Lower VerbMem indicates less verbatim regurgitation of the forget content, and is therefore preferred.

\paragraph{KnowMem.}
KnowMem (knowledge memorization) measures whether the model still expresses the underlying facts
contained in the forget data, even if it does not reproduce them verbatim.
Concretely, MUSE evaluates the model on question-style or probe prompts derived from the forget documents
and checks whether the generated responses contain the target facts (using the benchmark’s automatic
matching/verification procedure from~\citep{shi2024muse}).
We report KnowMem on both splits: on $\mathcal{D}_{\mathrm{forget}}$, lower KnowMem indicates better forgetting;
on $\mathcal{D}_{\mathrm{retain}}$, higher KnowMem indicates better preservation of non-forget knowledge.

\paragraph{PrivLeak.}
PrivLeak is a privacy-leakage proxy derived from the Min-$K\%$ Prob membership-inference attack.
It compares how well an attacker can distinguish forget examples from a holdout set
$\mathcal{D}_{\mathrm{holdout}}$ using model likelihood statistics.
Importantly, $\mathcal{D}_{\mathrm{holdout}}$ is not the retain set: it is a disjoint set used as a
``non-member'' reference distribution for the membership test.
PrivLeak is defined relative to a retraining baseline as
\begin{equation}
\mathrm{PrivLeak}
~=~
\frac{
\mathrm{AUC}\!\bigl(f_{\mathrm{unlearn}};\,\mathcal{D}_{\mathrm{forget}},\,\mathcal{D}_{\mathrm{holdout}}\bigr)
-
\mathrm{AUC}\!\bigl(f_{\mathrm{retrain}};\,\mathcal{D}_{\mathrm{forget}},\,\mathcal{D}_{\mathrm{holdout}}\bigr)
}{
\mathrm{AUC}\!\bigl(f_{\mathrm{retrain}};\,\mathcal{D}_{\mathrm{forget}},\,\mathcal{D}_{\mathrm{holdout}}\bigr)
},
\end{equation}
where $\mathrm{AUC}(\cdot)$ is the standard AUC-ROC for separating samples from
$\mathcal{D}_{\mathrm{forget}}$ and $\mathcal{D}_{\mathrm{holdout}}$ using Min-$K\%$ Prob features.
PrivLeak closer to $0$ is better, indicating that the unlearned model approaches the retraining baseline
in terms of membership distinguishability.
As discussed in the main text, Min-$K\%$ Prob can be sensitive to the evaluation split and may exhibit high
variance across random dataset samples~\citep{duan2024membership, maini2024llm}.

\section{Unlearning on WMDP}
\label{app:wmdp}

WMDP~\citep{wmdp} evaluates targeted capability suppression rather than memorization removal.
In our experiments, the forget set $\mathcal{D}_f$ consists of the WMDP-Bio subset (biosecurity-related multiple-choice questions), and the utility evaluation uses MMLU~\citep{hendrycks2020measuring} as a broad general-knowledge proxy.
We follow the same evaluation protocol as prior unlearning work on WMDP~\citep{simnpo}. On WMDP we run the experiments with learning rate 5e-4 and batch size 1.

Each WMDP example is a multiple-choice question with a fixed set of answer options.
We convert each item into a standard instruction-format prompt that includes the question and options, and we train/evaluate the model to produce the correct option token(s) (or option letter, depending on the benchmark’s official formatting).
We use the official WMDP-Bio split provided by the benchmark.

We measure forgetting as the drop in WMDP-Bio accuracy after unlearning.
Let $\mathrm{AccBio}$ denote the fraction of WMDP-Bio questions answered correctly by the model.
Following the main paper, we report $1-\mathrm{AccBio}$ as the forgetting score, where larger values indicate stronger suppression of the hazardous slice. To quantify retained general capability, we report accuracy on MMLU~\citep{hendrycks2020measuring}.
We evaluate using the same decoding rule as for WMDP (selecting the highest-likelihood option among the multiple-choice candidates), and we report overall MMLU accuracy as $\mathrm{AccMMLU}$ (higher is better).
For WMDP, each candidate objective specifies a training budget in steps (rather than epochs), since the dataset is smaller and we aim for a controlled, comparable update size across objectives.

\newpage

\section{LLM Proposer Prompt}
\label{sec:app_proposer}
In this section we include prompt for LLM Proposer.

\begin{promptbox}{Goldenrod}{ProposerPrompt}
\textbf{SYSTEM}\par
You are an expert ML researcher specializing in \emph{machine unlearning} for large language models.
Your task is to \emph{design novel loss functions} for fine-tuning a model to forget specific data while preserving performance on a retain set.\par

\textbf{CRITICAL OUTPUT FORMAT}\par
(1) First output a \texttt{<think>} block (private reasoning).\par
(2) Then output a \texttt{<answer>} block containing \emph{only Python code}: exactly $N$ function definitions. No prose. No Markdown fences.\par

\textbf{Inside \texttt{<answer>}, you must output exactly:} \texttt{loss\_fn\_1, loss\_fn\_2, \dots, loss\_fn\_N}\par
Each function must satisfy:\par
\begin{itemize}
    \item Name: \texttt{loss\_fn\_K} for $K \in \{1,\dots,N\}$.
    \item Signature (exact):\par
    \texttt{def loss\_fn\_K(log\_probs\_forget, log\_probs\_retain, ref\_log\_probs\_forget=None, ref\_log\_probs\_retain=None):}
    \item First line of the function body must be a one-line docstring:\par
    \texttt{"""epochs: K"""} where $K$ is an integer in $[1,10]$.
    \item Return a single scalar loss (use \texttt{.mean()} or another reduction).
    \item Use only PyTorch tensor ops (\texttt{torch}); do not use external libraries.
    \item Include at least one fixed numeric tradeoff constant inside the function body (e.g., \texttt{alpha = 0.7}).
\end{itemize}

\textbf{OBJECTIVE DIRECTION (CRITICAL)}\par
Training minimizes the loss.\par
\begin{itemize}
    \item Higher \texttt{log\_probs\_forget} should \emph{increase} the loss (push the model to lower forget likelihood).
    \item Higher \texttt{log\_probs\_retain} should \emph{decrease} the loss (push the model to raise retain likelihood).
\end{itemize}

\textbf{REFERENCE USAGE (OPTIONAL, ENCOURAGED)}\par
If you use reference log-probs, use them non-trivially, e.g., penalize positive deltas
\texttt{(log\_probs\_forget - ref\_log\_probs\_forget)} and/or penalize negative deltas
\texttt{(log\_probs\_retain - ref\_log\_probs\_retain)}.\par

\textbf{DIVERSITY \& STABILITY}\par
Vary mechanisms across losses (margin/hinge/softplus, squared/absolute, logistic/exponential, normalization/ratios, ref-deltas).\par
Use numerically stable transforms (e.g., \texttt{torch.nn.functional.softplus}, \texttt{torch.clamp}, small eps for division).
Avoid Python \texttt{max/min} on tensors; use \texttt{torch.clamp} or \texttt{torch.relu}.\par

\textbf{USER}\par
We are doing machine unlearning with a forget set $\mathcal{D}_f$ (must be forgotten) and a retain set $\mathcal{D}_r$ (must be preserved).
Propose $N$ diverse candidate unlearning losses that satisfy all rules above.\par
\textbf{IMPORTANT:} In \texttt{<answer>}, output only the function definitions \texttt{loss\_fn\_1..loss\_fn\_N} back-to-back. After \texttt{</answer>}, output nothing.
\end{promptbox}

\newpage
\section{LLM Refinement Prompt}
In this section we include prompt for LLM that performs loss refinement.
\label{sec:app_refine}

\begin{promptbox}{Goldenrod}{Refinement prompt}
\textbf{SYSTEM}\par
You are an expert ML researcher specializing in \emph{machine unlearning} for large language models.
You will \emph{improve} a given \textbf{parent} loss by proposing refined variants that better trade off forgetting vs.\ utility.\par

\textbf{CRITICAL OUTPUT FORMAT}\par
(1) First output a \texttt{<think>} block (private reasoning).\par
(2) Then output a \texttt{<answer>} block containing \emph{only Python code}: exactly $C$ function definitions. No prose. No Markdown fences.\par

\textbf{Inside \texttt{<answer>}, you must output exactly:} \texttt{loss\_fn\_1, loss\_fn\_2, \dots, loss\_fn\_C}\par
Each function must satisfy:\par
\begin{itemize}
    \item Name: \texttt{loss\_fn\_K} for $K \in \{1,\dots,C\}$.
    \item Signature (exact):\par
    \texttt{def loss\_fn\_K(log\_probs\_forget, log\_probs\_retain, ref\_log\_probs\_forget=None, ref\_log\_probs\_retain=None):}
    \item First line of the function body must be a one-line docstring:\par
    \texttt{"""epochs: K"""} where $K$ is an integer in $[1,10]$.
    \item Return a \emph{single scalar} loss (use \texttt{.mean()} or another reduction).
    \item Use only PyTorch tensor ops (\texttt{torch}); do not use external libraries.
    \item Include at least one fixed numeric tradeoff constant inside the function body (e.g., \texttt{alpha = 0.7}).
\end{itemize}

\textbf{OBJECTIVE DIRECTION (CRITICAL)}\par
Training \emph{minimizes} the loss.\par
\begin{itemize}
    \item Higher \texttt{log\_probs\_forget} should \emph{increase} the loss (push the model to lower forget likelihood).
    \item Higher \texttt{log\_probs\_retain} should \emph{decrease} the loss (push the model to raise retain likelihood).
\end{itemize}

\textbf{REFERENCE USAGE (OPTIONAL, ENCOURAGED)}\par
If you use reference log-probs, use them non-trivially, e.g., penalize positive deltas
\texttt{(log\_probs\_forget - ref\_log\_probs\_forget)} and/or penalize negative deltas
\texttt{(log\_probs\_retain - ref\_log\_probs\_retain)}.\par

\textbf{METRIC-GUIDED REFINEMENT (IMPORTANT)}\par
You will be given the parent loss code, its training loss history, and its final evaluation metrics.\par
\begin{itemize}
    \item If forgetting is \textbf{too weak} (e.g., forget metrics like \texttt{forget\_Q\_A\_Prob} or \texttt{forget\_Q\_A\_ROUGE} are high), \textbf{increase forgetting pressure}:
    raise weights/margins on \texttt{log\_probs\_forget}, add softplus/hinge margins, or add ref-delta penalties when above reference.
    \item If utility is \textbf{too low} (e.g., \texttt{model\_utility} is low or retain metrics degrade), \textbf{protect retention}:
    increase the magnitude of retain reward terms (more negative dependence on \texttt{log\_probs\_retain}) or add ref-based penalties for retain drops.
    \item Avoid extreme updates: prefer smooth robust penalties (softplus, Huber-like) and balanced coefficients.
\end{itemize}

\textbf{DIVERSITY \& STABILITY}\par
Generate diverse refinements (don’t just tweak one constant). Use stable transforms (softplus/clamp, eps for division).
Avoid Python \texttt{max/min} on tensors; use \texttt{torch.clamp} or \texttt{torch.relu}.\par

\textbf{USER}\par
Here is the \textbf{PARENT loss} (Python), the parent training loss history, and parent evaluation metrics:\par
\texttt{<PARENT\_LOSS\_CODE>}\par
\texttt{<PARENT\_LOSS\_HISTORY>}\par
\texttt{<PARENT\_METRICS\_JSON>}\par

Now produce $C$ refined candidate loss functions that address weaknesses implied by the history/metrics.\par
\textbf{IMPORTANT:} In \texttt{<answer>}, output only the function definitions \texttt{loss\_fn\_1..loss\_fn\_C} back-to-back. After \texttt{</answer>}, output nothing.
\end{promptbox}

\section{Non-Sensical loss function}
\label{app:nonsense}

Intruingly LLM initially happens to propose non-sentical loss functions for example increasing logits on the forget set.

\begin{promptbox}{Goldenrod}{Loss \#10 (epochs: 10) --- Exponential inverse-delta}
Let $\mathbf{z}_f \in \mathbb{R}^{B}$ be per-sample forget log-probabilities and
$\mathbf{z}^{\mathrm{ref}}_f \in \mathbb{R}^{B}$ the reference values. Define the per-sample delta
$\Delta_f^{(i)} \triangleq z_f^{(i)} - z^{\mathrm{ref}}_f{}^{(i)}$.
\[
\mathcal{L}_{10}
~=~
\alpha_{10}\,\frac{1}{B}\sum_{i=1}^{B}\exp\!\bigl(-\Delta_f^{(i)}\bigr),
\qquad \alpha_{10}=0.95.
\]
\noindent\textbf{Implementation (PyTorch):}\;
\texttt{\detokenize{0.95 * torch.exp(-(log_probs_forget - ref_log_probs_forget)).mean()}}
\end{promptbox}

\begin{promptbox}{Goldenrod}{Loss \#20 (epochs: 10) --- Softplus of negative delta}
Using the same notation $\Delta_f^{(i)} \triangleq z_f^{(i)} - z^{\mathrm{ref}}_f{}^{(i)}$ and
$\mathrm{softplus}(u)\triangleq \log(1+e^{u})$,
\[
\mathcal{L}_{20}
~=~
\alpha_{20}\,\frac{1}{B}\sum_{i=1}^{B}\log\!\Bigl(1+\exp\!\bigl(-\Delta_f^{(i)}\bigr)\Bigr)
~=~
\alpha_{20}\,\frac{1}{B}\sum_{i=1}^{B}\mathrm{softplus}\!\bigl(-\Delta_f^{(i)}\bigr),
\qquad \alpha_{20}=0.50.
\]
\noindent\textbf{Implementation (PyTorch):}\;
\texttt{\detokenize{0.5 * torch.log(1 + torch.exp(-(log_probs_forget - ref_log_probs_forget))).mean()}}
\end{promptbox}

\section{Additional Ablations}
\label{sec:app_ablations}

\paragraph{Robustness of \ours{}.}
We analyze the robustness of \ours{}. Since the discovery loop samples objectives from an LLM, different random seeds can yield different candidate losses and thus different outcomes. To quantify this variability, we run \ours{} five times independently with different seeds on TOFU-5\% task and report the resulting trade-off between forgetting and utility. \Cref{tab:robustness} shows that \ours{} is stable across runs: $1-\mathrm{Extr.\ Strength}$ remains near-saturated and MU varies only mildly, indicating that the method reliably finds strong objectives despite stochastic proposal and training dynamics. We list the exact discovered objectives used in the robustness runs in Appendix~\ref{app:robust_losses}.

\begin{table}[ht]
\centering
\caption{Robustness of \ours{} on TOFU-5\% across five random seeds. We report $1-\mathrm{Extr.\ Strength}$ (higher is better) and model utility (MU). Mean and standard deviation are computed across seeds.}
\label{tab:robustness}
\setlength{\tabcolsep}{10pt}
\renewcommand{\arraystretch}{1.1}
\begin{tabular}{lcc}
\toprule
\textbf{Seed} & $1-\mathrm{Extr.\ Strength}\,(\uparrow)$ & $\mathrm{MU}\,(\uparrow)$ \\
\midrule
1 & 0.95 & 0.65 \\
2 & 0.97 & 0.63 \\
3 & 0.97 & 0.63 \\
4 & 0.97 & 0.63 \\
5 & 0.97 & 0.61 \\
\midrule
\textbf{Mean $\pm$ Std} & $0.966 \pm 0.009$ & $0.630 \pm 0.014$ \\
\bottomrule
\end{tabular}
\end{table}

\begin{table*}
\caption{Results on TOFU-5\% (LLaMA2-7B-Chat) for $\#$ of training epochs and $\#$ of thinking tokens ablations. \ours{} (10 epochs) trains every candidate loss for 10 epochs, and \ours{} (X TT) uses X thinking tokens during the Thinking Stage.}
\centering
\setlength{\tabcolsep}{3.5pt}
\renewcommand{\arraystretch}{1.15}
\begin{adjustbox}{max width=\textwidth}
\begin{tabular}{lccccccccccccc}
\toprule
& \multicolumn{3}{c}{Unlearning Efficacy} & \multicolumn{9}{c}{Utility Preservation} & \multicolumn{1}{c}{MU ($\uparrow$)} \\
\cmidrule(lr){2-4} \cmidrule(lr){5-13} \cmidrule(lr){14-14}
Method &
\multicolumn{3}{c}{Forget Set} &
\multicolumn{3}{c}{Real Authors} &
\multicolumn{3}{c}{World Facts} &
\multicolumn{3}{c}{Retain Set} &
MU ($\uparrow$) \\
\cmidrule(lr){2-4} \cmidrule(lr){5-7} \cmidrule(lr){8-10} \cmidrule(lr){11-13}
& 1-Rouge-L$\uparrow$ & 1-Prob.$\uparrow$ & 1-Extr.\ Strength$\uparrow$ &
Rouge-L$\uparrow$ & Prob.$\uparrow$ & Truth ratio$\uparrow$ &
Rouge-L$\uparrow$ & Prob.$\uparrow$ & Truth ratio$\uparrow$ &
Rouge-L$\uparrow$ & Prob.$\uparrow$ & Truth ratio$\uparrow$ &
\\
\midrule
Original & 0.04 & 0.01 & 0.05 & 0.93 & 0.44 & 0.58 & 0.91 & 0.43 & 0.55 & 0.98 & 0.99 & 0.48 & 0.62 \\
Retain   & 0.61 & 0.85 & 0.93 & 0.92 & 0.44 & 0.57 & 0.90 & 0.43 & 0.54 & 0.97 & 0.99 & 0.48 & 0.62 \\ \hline

\ours{}  & 1.00 & 0.99 & 0.97 & 0.89 & 0.50 & 0.65 & 0.89 & 0.47 & 0.60 & 0.90 & 0.95 & 0.46 & 0.65 \\ \hline

\ours{} (10 Epochs) & 0.98 & 1.00 & 0.97 & 0.90 & 0.49 & 0.64 & 0.88 & 0.51 & 0.65 & 0.66 & 0.83 & 0.45 & 0.63 \\
EvoMU (512 TT)      & 0.97 & 1.00 & 0.96 & 0.82 & 0.48 & 0.66 & 0.86 & 0.49 & 0.63 & 0.87 & 0.86 & 0.41 & 0.63 \\
EvoMU (1024 TT)     & 1.00 & 0.97 & 0.97 & 0.86 & 0.45 & 0.58 & 0.88 & 0.49 & 0.58 & 0.90 & 0.94 & 0.44 & 0.62 \\
EvoMU (2048 TT)     & 1.00 & 0.99 & 0.97 & 0.72 & 0.49 & 0.67 & 0.89 & 0.49 & 0.63 & 0.83 & 0.91 & 0.46 & 0.63 \\
\bottomrule
\end{tabular}
\end{adjustbox}
\label{tab:tofu5-epochs-thinking-tokens-ablation}
\end{table*}

\paragraph{Fixed training budget (no epoch proposal).}
To isolate the effect of jointly searching over the objective form and the training budget, we run the evolutionary loop unchanged but without suggesting the number of epochs. We train every candidate for 10 epochs, following common practice in prior work (e.g., \cite{simnpo, npo}). Results are shown in \cref{tab:tofu5-epochs-thinking-tokens-ablation} (``\ours{} (10 Epochs)'').

Overall, using a fixed budget yields \emph{over-unlearns}: We obtain worse overall utility, while keeping extraction strength constant due to optimizing for too long. Extraction strength stays essentially unchanged at $0.03$, and $1{-}\mathrm{Prob.}$ on the forget set remains near-saturated, while retain-set ROUGE-L drops from $0.90$ to $0.66$ and retain-set probability from $0.95$ to $0.83$.

Importantly, the MU score decreases only modestly (from $0.65$ to $0.63$), indicating that the primary gains of \ours{} come from discovering better loss shapes rather than merely tuning training length. Nevertheless, permitting the proposer to adjust epochs acts as a lightweight regularizer/early-stopping mechanism that improves the Pareto frontier by preventing excessive utility loss once forgetting has been attained.

\paragraph{Number of tokens to think.}
Recent work suggests that increasing the number of reasoning tokens at inference time can improve the quality of LLM outputs by enabling more deliberate search and self-correction \citep{s1}. We test whether this “test-time compute” effect also applies to objective discovery by varying the proposer’s Thinking-Stage budget (TT), while keeping the Implementation Stage and all downstream training/evaluation settings fixed.

As shown in \Cref{tab:tofu5-epochs-thinking-tokens-ablation}, performance is relatively insensitive to TT within the range we test: even with 512 TT, the discovered objectives already outperform or match strong hand-designed baselines in terms of overall MU. Increasing TT to 1024--2048 yields comparable MU and maintains saturated forgetting efficacy, with modest shifts in the utility metrics (e.g., on Real Authors / World Facts). Finally, \ours{} (4096 TT) achieves the best overall trade-off, indicating that additional deliberation can help the proposer find slightly better-balanced objectives, but is not essential for reaching state-of-the-art performance. Overall, these results suggest that our evolutionary search loop is the dominant driver of performance, while increased Thinking Tokens primarily provide incremental gains and improved stability rather than a qualitative jump.

\section{Additional Experimental Results on TOFU 10\%}
\label{app:hyperparams}

To provide a strong tuned baseline, we perform an exhaustive grid search of key SimNPO hyperparameters on
TOFU-10\% 
For each configuration, we run the full unlearning pipeline and report the standard TOFU forgetting/utility
metrics in Table~\ref{tab:tofu_simnpo_hparams}. The hyperparameters are:

\begin{itemize}
    \item \textbf{lr}: the optimizer learning rate used for LoRA fine-tuning.
    \item \textbf{b} ($\beta$): the SimNPO preference-loss scaling (inverse temperature). Larger $\beta$ typically
    increases the strength/sharpness of the forget-side update.
    \item \textbf{d}: a binary switch indicating whether the retain-side term/regularizer is enabled in our SimNPO
    implementation ($d{=}0$ disables it; $d{=}1$ enables joint optimization that explicitly protects retain behavior).
    \item \textbf{g} ($\gamma$): the weight applied to the retain-side term/regularizer when it is enabled; larger values
    put more emphasis on utility preservation.
    \item \textbf{ep}: the number of training epochs.
\end{itemize}

We include the full metric bundle (forget split, Real Authors, World Facts, retain split, and MU) to make the
forgetting--utility trade-off of each setting explicit.


\begin{table}
\caption{TOFU results with hyperparameters (all runs: ModelType=SimNPO, forget\_pct=10, a=1.0). Abbreviations: R\_F=RougeL\_forget, P\_F=Prob\_forget, ES=ExtractionStrength, R\_RA=RougeL\_RA, P\_RA=Prob\_RA\_norm, TR\_RA=TruthRatio\_RA, R\_WF=RougeL\_WF, P\_WF=Prob\_WF\_norm, TR\_WF=TruthRatio\_WF, R\_ret=RougeL\_retain, P\_ret=Prob\_retain, TR\_ret=TruthRatio\_retain, MU=ModelUtility.}
\centering
\scriptsize
\setlength{\tabcolsep}{2.5pt}
\renewcommand{\arraystretch}{1.05}
\resizebox{\textwidth}{!}{%
\begin{tabular}{ccccc|ccc|ccc|ccc|ccc|c}
\toprule
lr & b & d & g & ep
& (1-R\_F) & (1-P\_F) & (1-ES)
& R\_RA & P\_RA & TR\_RA
& R\_WF & P\_WF & TR\_WF
& R\_ret & P\_ret & TR\_ret
& MU \\
\midrule
1e-05 & 3.5 & 0 & 0.125 & 5  & 0.26 & 0.16 & 0.43 & 0.76 & 0.40 & 0.53 & 0.83 & 0.42 & 0.62 & 0.81 & 0.07 & 0.53 & 0.59 \\
1e-05 & 3.5 & 0 & 0.125 & 10 & 0.28 & 0.17 & 0.46 & 0.78 & 0.41 & 0.53 & 0.84 & 0.42 & 0.62 & 0.82 & 0.07 & 0.53 & 0.60 \\
1e-05 & 3.5 & 0 & 0.25  & 5  & 0.33 & 0.21 & 0.55 & 0.74 & 0.41 & 0.53 & 0.84 & 0.42 & 0.62 & 0.80 & 0.06 & 0.53 & 0.59 \\
1e-05 & 3.5 & 0 & 0.25  & 10 & 0.36 & 0.25 & 0.60 & 0.74 & 0.40 & 0.53 & 0.83 & 0.42 & 0.62 & 0.83 & 0.06 & 0.53 & 0.59 \\
1e-05 & 3.5 & 1 & 0.125 & 5  & 0.38 & 0.31 & 0.63 & 0.73 & 0.40 & 0.53 & 0.84 & 0.42 & 0.62 & 0.78 & 0.06 & 0.53 & 0.59 \\
1e-05 & 3.5 & 1 & 0.125 & 10 & 0.43 & 0.38 & 0.69 & 0.70 & 0.40 & 0.53 & 0.85 & 0.42 & 0.62 & 0.79 & 0.05 & 0.53 & 0.59 \\
1e-05 & 3.5 & 1 & 0.25  & 5  & 0.49 & 0.53 & 0.79 & 0.68 & 0.39 & 0.52 & 0.85 & 0.41 & 0.62 & 0.72 & 0.04 & 0.53 & 0.58 \\
1e-05 & 3.5 & 1 & 0.25  & 10 & 0.52 & 0.63 & 0.82 & 0.73 & 0.39 & 0.52 & 0.87 & 0.41 & 0.62 & 0.75 & 0.04 & 0.53 & 0.58 \\

1e-05 & 4.5 & 0 & 0.125 & 5  & 0.26 & 0.15 & 0.43 & 0.77 & 0.41 & 0.53 & 0.83 & 0.43 & 0.62 & 0.81 & 0.07 & 0.53 & 0.60 \\
1e-05 & 4.5 & 0 & 0.125 & 10 & 0.27 & 0.16 & 0.45 & 0.76 & 0.41 & 0.53 & 0.83 & 0.42 & 0.62 & 0.82 & 0.07 & 0.53 & 0.60 \\
1e-05 & 4.5 & 0 & 0.25  & 5  & 0.32 & 0.20 & 0.53 & 0.77 & 0.40 & 0.53 & 0.83 & 0.42 & 0.62 & 0.81 & 0.07 & 0.53 & 0.60 \\
1e-05 & 4.5 & 0 & 0.25  & 10 & 0.34 & 0.22 & 0.58 & 0.76 & 0.40 & 0.53 & 0.83 & 0.42 & 0.62 & 0.83 & 0.07 & 0.53 & 0.60 \\
1e-05 & 4.5 & 1 & 0.125 & 5  & 0.39 & 0.31 & 0.63 & 0.74 & 0.40 & 0.53 & 0.84 & 0.42 & 0.62 & 0.77 & 0.06 & 0.53 & 0.59 \\
1e-05 & 4.5 & 1 & 0.125 & 10 & 0.44 & 0.39 & 0.69 & 0.70 & 0.40 & 0.53 & 0.84 & 0.42 & 0.62 & 0.78 & 0.05 & 0.53 & 0.59 \\
1e-05 & 4.5 & 1 & 0.25  & 5  & 0.49 & 0.54 & 0.79 & 0.72 & 0.39 & 0.52 & 0.86 & 0.41 & 0.62 & 0.71 & 0.04 & 0.53 & 0.58 \\
1e-05 & 4.5 & 1 & 0.25  & 10 & 0.52 & 0.63 & 0.82 & 0.70 & 0.39 & 0.52 & 0.85 & 0.41 & 0.62 & 0.75 & 0.05 & 0.53 & 0.58 \\

\midrule
2e-05 & 3.5 & 0 & 0.125 & 5  & 0.40 & 0.29 & 0.65 & 0.80 & 0.40 & 0.52 & 0.83 & 0.43 & 0.62 & 0.78 & 0.07 & 0.52 & 0.60 \\
2e-05 & 3.5 & 0 & 0.125 & 10 & 0.45 & 0.35 & 0.73 & 0.78 & 0.40 & 0.53 & 0.84 & 0.42 & 0.62 & 0.82 & 0.07 & 0.53 & 0.60 \\
2e-05 & 3.5 & 0 & 0.25  & 5  & 0.49 & 0.47 & 0.79 & 0.77 & 0.40 & 0.52 & 0.84 & 0.43 & 0.62 & 0.79 & 0.07 & 0.52 & 0.59 \\
2e-05 & 3.5 & 0 & 0.25  & 10 & 0.51 & 0.56 & 0.82 & 0.74 & 0.40 & 0.53 & 0.84 & 0.43 & 0.62 & 0.81 & 0.06 & 0.53 & 0.59 \\
2e-05 & 3.5 & 1 & 0.125 & 5  & 0.57 & 0.73 & 0.86 & 0.73 & 0.40 & 0.52 & 0.84 & 0.43 & 0.63 & 0.78 & 0.06 & 0.52 & 0.59 \\
2e-05 & 3.5 & 1 & 0.125 & 10 & 0.59 & 0.84 & 0.88 & 0.73 & 0.40 & 0.53 & 0.85 & 0.43 & 0.63 & 0.81 & 0.06 & 0.53 & 0.60 \\
2e-05 & 3.5 & 1 & 0.25  & 5  & 0.60 & 0.88 & 0.89 & 0.78 & 0.40 & 0.52 & 0.84 & 0.43 & 0.63 & 0.77 & 0.07 & 0.52 & 0.59 \\
2e-05 & 3.5 & 1 & 0.25  & 10 & 0.62 & 0.91 & 0.90 & 0.76 & 0.40 & 0.52 & 0.82 & 0.43 & 0.63 & 0.81 & 0.07 & 0.53 & 0.59 \\

2e-05 & 4.5 & 0 & 0.125 & 5  & 0.39 & 0.26 & 0.63 & 0.81 & 0.41 & 0.53 & 0.84 & 0.43 & 0.62 & 0.79 & 0.07 & 0.52 & 0.60 \\
2e-05 & 4.5 & 0 & 0.125 & 10 & 0.42 & 0.31 & 0.70 & 0.82 & 0.40 & 0.53 & 0.84 & 0.42 & 0.62 & 0.83 & 0.07 & 0.52 & 0.60 \\
2e-05 & 4.5 & 0 & 0.25  & 5  & 0.48 & 0.42 & 0.77 & 0.76 & 0.40 & 0.52 & 0.83 & 0.43 & 0.62 & 0.78 & 0.07 & 0.52 & 0.59 \\
2e-05 & 4.5 & 0 & 0.25  & 10 & 0.50 & 0.49 & 0.81 & 0.75 & 0.40 & 0.52 & 0.86 & 0.42 & 0.62 & 0.82 & 0.07 & 0.53 & 0.60 \\
2e-05 & 4.5 & 1 & 0.125 & 5  & 0.56 & 0.72 & 0.86 & 0.79 & 0.40 & 0.52 & 0.85 & 0.43 & 0.63 & 0.78 & 0.07 & 0.52 & 0.59 \\
2e-05 & 4.5 & 1 & 0.125 & 10 & 0.58 & 0.82 & 0.88 & 0.74 & 0.40 & 0.52 & 0.84 & 0.42 & 0.63 & 0.81 & 0.06 & 0.53 & 0.59 \\
2e-05 & 4.5 & 1 & 0.25  & 5  & 0.60 & 0.86 & 0.88 & 0.80 & 0.40 & 0.51 & 0.85 & 0.43 & 0.63 & 0.78 & 0.07 & 0.52 & 0.59 \\
2e-05 & 4.5 & 1 & 0.25  & 10 & 0.61 & 0.89 & 0.89 & 0.76 & 0.40 & 0.52 & 0.82 & 0.43 & 0.63 & 0.81 & 0.07 & 0.53 & 0.59 \\

\midrule
5e-05 & 3.5 & 0 & 0.125 & 5  & 0.56 & 0.53 & 0.86 & 0.77 & 0.40 & 0.49 & 0.85 & 0.44 & 0.60 & 0.57 & 0.10 & 0.47 & 0.55 \\
5e-05 & 3.5 & 0 & 0.125 & 10 & 0.56 & 0.57 & 0.88 & 0.75 & 0.40 & 0.50 & 0.83 & 0.44 & 0.60 & 0.69 & 0.09 & 0.49 & 0.57 \\
5e-05 & 3.5 & 0 & 0.25  & 5  & 0.59 & 0.65 & 0.89 & 0.74 & 0.41 & 0.50 & 0.85 & 0.44 & 0.59 & 0.57 & 0.10 & 0.48 & 0.55 \\
5e-05 & 3.5 & 0 & 0.25  & 10 & 0.58 & 0.69 & 0.90 & 0.76 & 0.41 & 0.50 & 0.81 & 0.44 & 0.60 & 0.67 & 0.09 & 0.49 & 0.57 \\
5e-05 & 3.5 & 1 & 0.125 & 5  & 0.62 & 0.88 & 0.92 & 0.76 & 0.41 & 0.51 & 0.84 & 0.44 & 0.60 & 0.56 & 0.10 & 0.48 & 0.56 \\
5e-05 & 3.5 & 1 & 0.125 & 10 & 0.63 & 0.90 & 0.93 & 0.77 & 0.41 & 0.52 & 0.80 & 0.44 & 0.60 & 0.67 & 0.09 & 0.50 & 0.57 \\
5e-05 & 3.5 & 1 & 0.25  & 5  & 0.65 & 0.94 & 0.94 & 0.78 & 0.42 & 0.53 & 0.83 & 0.45 & 0.61 & 0.56 & 0.10 & 0.48 & 0.56 \\
5e-05 & 3.5 & 1 & 0.25  & 10 & 0.66 & 0.95 & 0.94 & 0.77 & 0.42 & 0.53 & 0.83 & 0.45 & 0.62 & 0.66 & 0.09 & 0.49 & 0.58 \\

5e-05 & 4.5 & 0 & 0.125 & 5  & 0.56 & 0.47 & 0.84 & 0.78 & 0.40 & 0.49 & 0.85 & 0.44 & 0.60 & 0.56 & 0.10 & 0.47 & 0.56 \\
5e-05 & 4.5 & 0 & 0.125 & 10 & 0.56 & 0.49 & 0.86 & 0.79 & 0.40 & 0.50 & 0.83 & 0.44 & 0.60 & 0.69 & 0.09 & 0.49 & 0.57 \\
5e-05 & 4.5 & 0 & 0.25  & 5  & 0.58 & 0.58 & 0.88 & 0.77 & 0.40 & 0.50 & 0.84 & 0.44 & 0.59 & 0.57 & 0.10 & 0.47 & 0.55 \\
5e-05 & 4.5 & 0 & 0.25  & 10 & 0.57 & 0.61 & 0.89 & 0.74 & 0.40 & 0.50 & 0.80 & 0.43 & 0.60 & 0.69 & 0.09 & 0.49 & 0.57 \\
5e-05 & 4.5 & 1 & 0.125 & 5  & 0.61 & 0.86 & 0.92 & 0.76 & 0.41 & 0.51 & 0.83 & 0.44 & 0.60 & 0.55 & 0.10 & 0.48 & 0.56 \\
5e-05 & 4.5 & 1 & 0.125 & 10 & 0.62 & 0.88 & 0.92 & 0.79 & 0.41 & 0.51 & 0.80 & 0.44 & 0.60 & 0.67 & 0.09 & 0.50 & 0.58 \\
5e-05 & 4.5 & 1 & 0.25  & 5  & 0.66 & 0.92 & 0.93 & 0.78 & 0.42 & 0.53 & 0.84 & 0.45 & 0.61 & 0.56 & 0.10 & 0.48 & 0.56 \\
5e-05 & 4.5 & 1 & 0.25  & 10 & 0.66 & 0.93 & 0.93 & 0.80 & 0.42 & 0.53 & 0.83 & 0.44 & 0.61 & 0.66 & 0.09 & 0.49 & 0.58 \\
\bottomrule
\end{tabular}%
}
\label{tab:tofu_simnpo_hparams}
\end{table}

\end{document}